%% file: Main.tex
\documentclass[journal]{IEEEtran}
\usepackage{makecell}
\usepackage{booktabs}
\ifCLASSINFOpdf
  \usepackage{graphicx}
  \usepackage{svg}
\else
  \usepackage[dvips]{graphicx}
\fi
\usepackage{amsmath,amsfonts,amssymb,adjustbox}
\usepackage{algorithm,algpseudocode,}

\usepackage{stfloats}
\usepackage{gensymb}
\usepackage{textcomp}
\usepackage{hyperref}

\begin{document}

\title{Zero-Shot Benchmarking of Unified Multimodal Models for Shallow-Water UAV Image Distortion Assessment and Restoration}

\author{Dario~Scilla$^{1}$,
        Omar~A.~Lopez$^{1}$,
        Mattia~Rosso$^{2}$,
        Kasper~Johansen$^{1}$,
        and Matthew~F.~McCabe$^{1}$%
\thanks{$^{1}$Biological and Environmental Science and Engineering Division, 
Climate and Livability Initiative, King Abdullah University of Science and Technology, 
Thuwal 23955-6900, Saudi Arabia (e-mail: dario.scilla@kaust.edu.sa; 
omar.lopezcamargo@kaust.edu.sa; kasper.johansen@kaust.edu.sa; matthew.mccabe@kaust.edu.sa).}%
\thanks{$^{2}$Computer, Electrical and Mathematical Sciences and Engineering, 
King Abdullah University of Science and Technology, Thuwal 23955-6900, Saudi Arabia
(mattia.rosso@kaust.edu.sa).}%
}

\maketitle

\begin{abstract}
Shallow-water UAV imagery provides centimeter-scale observations of benthic environments, but its interpretation is often degraded by water-induced distortions. Recent unified multimodal large language models (MLLMs), which combine image understanding and generation within a single architecture, offer a potential framework for both assessing and restoring these distortions. \emph{To our knowledge, we present the first systematic zero-shot benchmark of unified MLLMs that jointly assesses water-induced distortions and restores shallow-water imagery within the same model framework.} Five open-weight unified MLLMs are evaluated with frozen pretrained weights and without access to paired reference images or descriptions of the true seabed. The benchmark contains 1,000 paired images, comprising 600 semi-synthetic UAV scenes generated by applying synthetic water distortions to real shallow-water imagery and 400 fully synthetic through-water scenes, with three independent human annotations of five distortion types. Four assessment prompts (P0--P3), ranging from direct instruction to structured task decomposition, are evaluated across all 1,000 images. Restoration is then evaluated on a common 100-image subset using four diagnosis-guided restoration prompts (R0--R3). Assessment is strongly model- and prompt-dependent, with a best pooled presence balanced accuracy of 0.654 and systematic prediction biases across models. Restoration is substantially less reliable: the best pooled PSNR gain is only $+0.255$~dB, no unified MLLM improves pooled LPIPS, and some outputs introduce structural changes or hallucinated content. Two restoration-only baselines show similarly limited transfer, with InstructIR reaching only $+0.040$~dB PSNR improvement and PromptFix degrading all four pooled metrics under every prompt. These results show that recognizing shallow-water degradation does not imply reliable physical restoration and establish a reproducible baseline for future multimodal assessment and restoration methods.
\end{abstract}


%
\IEEEpeerreviewmaketitle

\section{Introduction}
\label{sec:introduction}

\IEEEPARstart{S}{hallow} coastal waters host some of the most productive and most threatened ecosystems on Earth \cite{Hobohm2021}. Coral reefs, seagrass meadows, and tidal flats support fisheries \cite{honda2013habitat}, buffer coastlines against storms \cite{guannel2016power}, sequester carbon \cite{kinsey1991significance}, and underpin the livelihoods of hundreds of millions of people, yet they are being degraded by climate change, sedimentation, and coastal development at an accelerating rate \cite{obura2019coral}. Timely, repeatable observation of these environments is therefore essential for conservation planning, restoration prioritization, and coastal management, which has driven sustained investment in remote sensing of the shallow seafloor \cite{gonzalez2020monitoring, kroon2014informing}.

Remote sensing is the primary tool for observing shallow-water environments at scale, but the right platform depends on the trade-off between coverage and resolution \cite{he2024remote}. Multispectral and hyperspectral satellites enable synoptic, repeat-pass mapping of coastal bathymetry and habitat over hundreds of kilometers \cite{duplancicleder2023optical}, while airborne bathymetric surveys trade coverage for higher spatial and vertical accuracy \cite{kromer2025sea}. Unmanned aerial vehicles (UAVs) close the remaining resolution gap: flying tens of meters above the water, they resolve individual coral colonies and seagrass patches at centimeter-to-decimeter scale that satellites and crewed aircraft cannot match, at a fraction of the cost and with on-demand revisit \cite{scilla2025computer}. This resolution advantage comes at the price of a much smaller footprint and, critically, a far greater sensitivity to the optical behavior of the water surface and column between the sensor and the seabed.

Because UAV imagery is acquired close to the water at high spatial resolution, it inherits distortions that coarser satellite pixels tend to average away \cite{david_structure_motion_2021}. Building on our own characterization of shallow-water UAV imagery \cite{scilla2025computer,kromer2025sea}, we consider five water-induced distortion types: \textit{(i) caustics}, focused bright networks projected onto the seabed by a wavy air--water interface \cite{scilla2025computer}; \textit{(ii) sun glint}, the specular reflection of the sun toward the sensor \cite{hedley2005sunglint}; \textit{(iii) refraction}, the geometric displacement of submerged features at the air--water boundary \cite{agrafiotis2020correcting}; \textit{(iv) turbidity}, the loss of contrast from light scattered by suspended sediment and organic matter \cite{cui2022water}; and \textit{(v) light absorption}, the wavelength-dependent attenuation that progressively removes color and detail with depth \cite{ackleson2003light}. Any UAV-based shallow-water application must contend with some combination of these five effects before the imagery is usable.

Correcting these effects was first approached with physics-based computer vision. Image formation models borrowed from atmospheric dehazing, such as the dark channel prior \cite{he2011dark}, and revised underwater-specific formulations that separately model backscatter and signal attenuation, such as Sea-thru \cite{akkaynak2019seathru}, recover a physically plausible clean image by inverting an explicit optical model. These methods are interpretable and require no training data, but they rely on assumptions (homogeneous water, a single dominant distortion type, known depth or range) that rarely hold across the heterogeneous, multi-distortion scenes typical of shallow reefs.

Deep learning relaxed these assumptions by learning the correction directly from data. Convolutional and generative-adversarial architectures such as FUnIE-GAN \cite{islam2020funie} learn to map degraded underwater images to enhanced ones end to end and have been shown, across surveys of the field, to outperform hand-crafted priors on real and synthetic benchmarks \cite{anwar2020diving}. However, most learned restorers are trained for a single degradation regime and a fixed output style, and their performance can degrade sharply outside the distribution of their training data, a persistent limitation for the varied optical conditions found in shallow coastal waters.

Generative diffusion models have since become the dominant paradigm for high-fidelity image restoration. Denoising diffusion probabilistic models \cite{ho2020ddpm} and their efficient latent-space formulation underlying Stable Diffusion \cite{rombach2022ldm} learn to invert a gradual noising process and can synthesize realistic detail that discriminative networks cannot, while more recent formulations such as ResShift \cite{yue2023resshift} recast restoration as a residual-shifting diffusion process to cut sampling cost by an order of magnitude. Diffusion models are now being adapted directly to the underwater domain, for instance to enhance turbid and color-degraded scenes with physics-aware attention mechanisms \cite{zhang2024underwaterdiffusion}, underscoring a broader shift from discriminative correction toward generative restoration.

Multimodal large language models (MLLMs) add a capability that none of the above methods have: the ability to reason about and describe an image's degradation in natural language before acting on it. Vision-language systems such as GPT-4V \cite{openai2023gpt4v} and LLaVA \cite{liu2023llava} exhibit in-context learning, adapting their behavior to a task from instructions and examples in the prompt alone, without retraining \cite{dong2023icl}. This has motivated recent work that prompts MLLMs directly for image quality assessment \cite{wu2024mllmiqa}, and has exposed a complication: MLLM outputs are highly sensitive to how a prompt is phrased and structured \cite{sahoo2024prompt,ismithdeen2025promptception}. Chain-of-thought and zero-shot reasoning prompts can partially stabilize this sensitivity \cite{wei2022chain,kojima2022large}. More recently, benchmarks such as MME-Unify, GIR-Bench, and RealUnify have begun to evaluate whether unified models can effectively connect their understanding and generation capabilities \cite{xie2026mmeunify,li2026girbench,shi2026realunify}. Their results reveal persistent limitations in transferring reasoning or understanding into faithful generation, despite both capabilities being integrated within the same model. Whether this coupling is sufficient for scientific image restoration under a strictly zero-shot, reference-free protocol remains largely unexplored, particularly for shallow-water UAV imagery.

Recent work has begun to explore related directions in underwater imaging. UQ-Bench evaluates MLLMs for underwater image-quality perception, including distortion recognition, free-text description, and quality scoring under a zero-shot setting \cite{cao2026uq}. However, its scope is limited to assessment and does not investigate whether the same models can generate restored imagery. In parallel, degradation-aware conditioning has been introduced for underwater image enhancement. CPLDP uses learned prompts representing color casts, uneven illumination, and haze to guide a task-specific enhancement network \cite{yao2026conditional}, while DIE2UIE extracts degradation- and semantic-aware visual-language cues to adaptively guide underwater enhancement \cite{wang2026degradation}. These approaches demonstrate the value of explicitly connecting degradation perception with enhancement, but rely on task-specific trained restoration frameworks rather than the pretrained generative capability of a unified MLLM.

This gap motivates our work. We present the first zero-shot benchmark of unified MLLMs for both the assessment and restoration of shallow-water UAV imagery within a single locally hosted model, without delegating diagnosis or synthesis to a separate network. Every experiment in this study is conducted strictly zero-shot and without reference supervision: all five models are evaluated with their pretrained weights kept entirely frozen, no fine-tuning, parameter-efficient adaptation, or in-context exemplars are used, and no model is ever shown the paired reference image, a restoration example, or any description of the true seabed, either in its prompt or otherwise. Each prediction is produced from a single distorted image and a text instruction alone; the reference image is used only afterward, by us, to score the outputs. Our contributions are the following: \emph{(i)} We curate and annotate a 1,000-image benchmark spanning two complementary sources: 600 semi-synthetic UAV scenes obtained by applying synthetic water distortions to real ground-truth imagery (KAUSTics dataset \cite{scilla2026kaustics}), and 400 fully synthetic through-water scenes (Sea-Undistort dataset \cite{kromer2025sea}), with three independent expert ratings of all five distortion types per image. \emph{(ii)} We design four progressively structured, zero-shot prompts for assessment and four for restoration, spanning direct, role-conditioned, definition-guided, and reasoning-guided variants, none of which reference ground truth, to systematically quantify how prompt design alone affects both distortion assessment and restoration quality. \emph{(iii)} We benchmark five open-weight unified MLLMs across both tasks under matched, frozen-weight conditions, reporting agreement-aware assessment accuracy, full-reference restoration quality, and computational cost. \emph{(iv)} By holding every model's weights fixed and withholding reference imagery throughout, we test whether the ability to assess shallow-water degradation translates into reliable image restoration using pretrained knowledge and prompting alone. This establishes a reproducible zero-shot baseline and exposes the limitations of general-purpose unified generative models for quantitative remote-sensing restoration.

\section{Methodology}
\label{sec:methodology}

\begin{figure}[!h]
\centering
\includegraphics[width=\columnwidth]{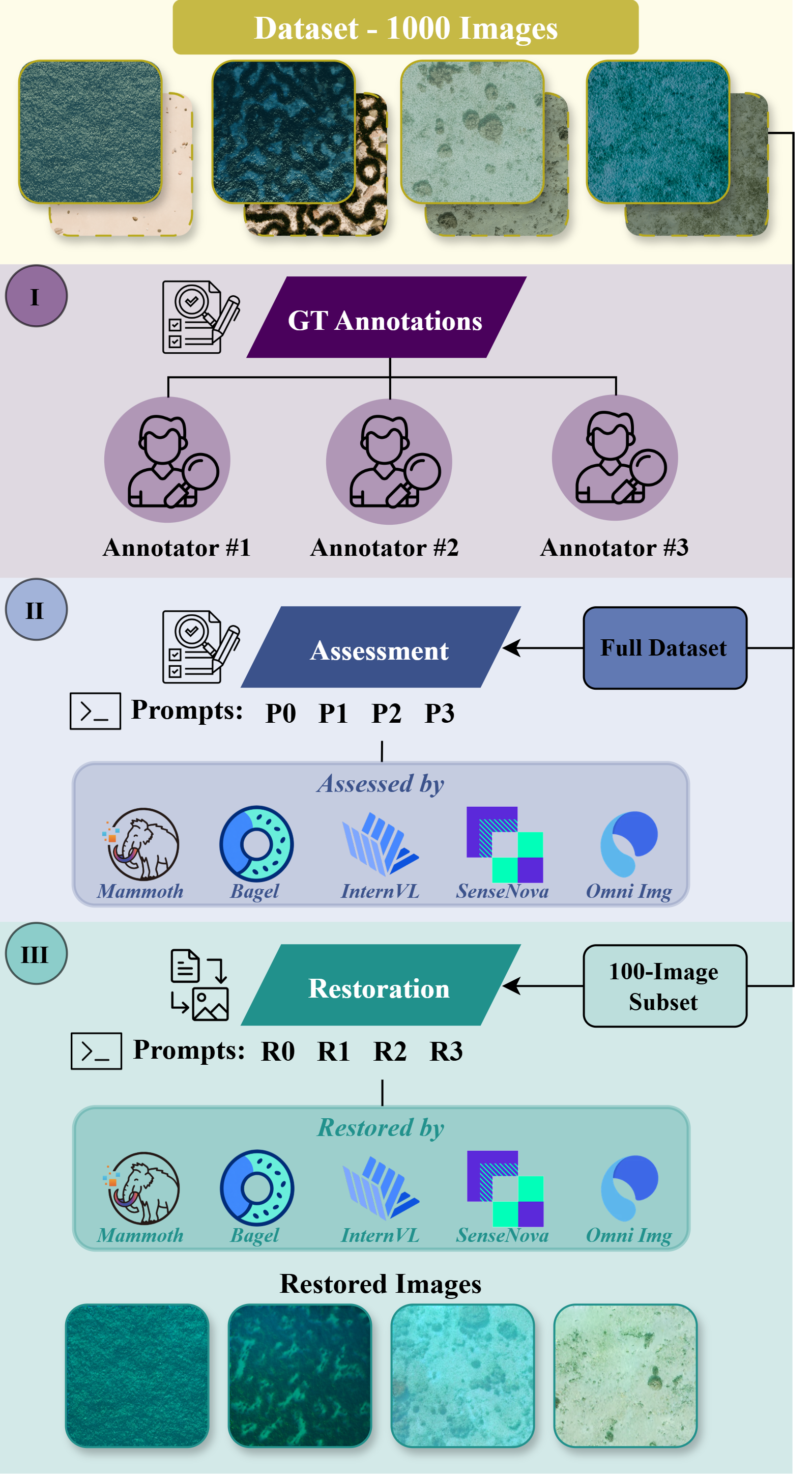}
\caption{Overview of the proposed benchmarking framework. \emph{(I)}: The 1,000-image dataset is first independently annotated by three human annotators for five water distortions. \emph{(II)}: Five unified MLLMs then assess all images using four zero-shot assessment prompts (P0--P3). \emph{(III)}: Model-generated diagnoses are then used to guide restoration on a 100-image subset using four restoration prompts (R0--R3).}
\label{fig:method}
\end{figure}

The proposed benchmarking framework evaluates whether unified multimodal large language models can both identify and correct water-induced distortions in shallow-water UAV imagery under a strictly zero-shot setting. An overview of the workflow is shown in Fig.~\ref{fig:method}. The benchmark contains 1,000 paired distorted/reference images obtained from two complementary datasets. All distorted images were independently annotated by three human annotators for the presence and severity of five water-induced distortions. The complete 1,000-image set was then used to evaluate the assessment capabilities of five unified MLLMs using four assessment prompts (P0--P3). For restoration, a 100-image subset was processed using four restoration prompts (R0--R3). The model-generated distortion assessments were converted into an ordered restoration plan according to the predicted severity, confidence, and a predefined distortion priority. Importantly, only the distorted images were provided to the annotators and models; paired reference images were withheld throughout inference and used only afterward.

\subsection{Dataset Preparation}
\label{subsec:dataset_preparation}

\begin{figure}[!h]
\centering
\includegraphics[width=\columnwidth]{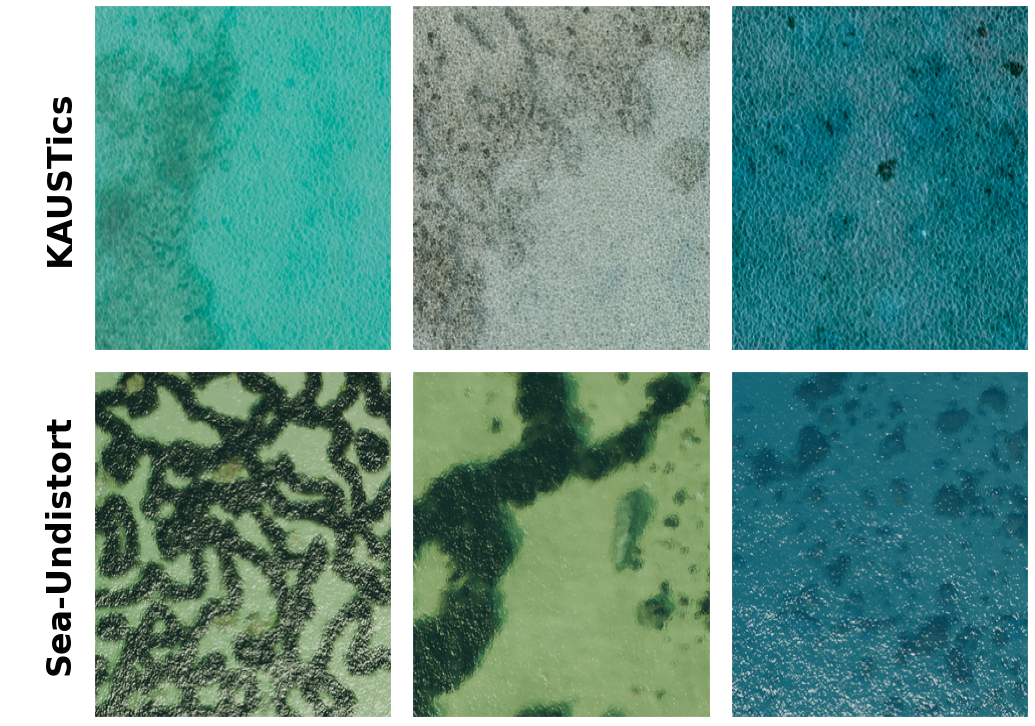}
\caption{Representative distorted images from the two source datasets. Top row: KAUSTics UAV semi-synthetic scenes ($1000\!\times\!1000$ pixels). Bottom row: Sea-Undistort synthetic through-water scenes ($512\!\times\!512$ pixels). Both sources show distortion patterns characteristic of shallow-water imaging, alongside variation in scene appearance and distortion severity.}
\label{fig:dataset_examples}
\end{figure}

The benchmark combines 1,000 paired distorted/reference shallow-water images from two complementary datasets. Sea-Undistort contributes 400 fully synthetic through-water scenes designed for paired image-restoration research \cite{kromer2025sea}, for which the $512\!\times\!512$ pixel products were used. KAUSTics contributes 600 semi-synthetic UAV scenes at $1000\!\times\!1000$ pixels, obtained by applying synthetic water-induced distortions to real ground-truth UAV imagery acquired during a shallow Red Sea reef survey \cite{scilla2026kaustics}. The distinction between the two datasets is therefore important: KAUSTics uses real UAV imagery as the undistorted reference and synthetically introduces the water effects, whereas both the scene generation and through-water image formation in Sea-Undistort are synthetic. Images were retained at their native resolutions to preserve the characteristics of each dataset. Representative examples are shown in Fig.~\ref{fig:dataset_examples}. The complete 1,000-image dataset was used for the assessment experiments. Restoration experiments were restricted to a common 100-image subset shared across all models because image generation is more computationally demanding than text-based assessment.

\subsection{Human Annotation Protocol}
\label{subsec:annotation_protocol} 

Because the visual severity of water-induced distortions is not always unambiguous, particularly for effects such as turbidity and refraction, the benchmark does not rely on a single human label. Instead,three human annotators independently evaluated every distorted image. The annotation panel comprised three Ph.D. students: two in Environmental Science and Engineering whose research involves remote sensing, and one in materials science/physics. For each image, the annotators assessed five distortion types: caustics, refraction, sun glint, turbidity, and light absorption. Each distortion was assigned one of four ordered severity levels: \textit{none}, \textit{low}, \textit{medium}, or \textit{high}. The annotation protocol required a severity selection for all five distortions before a record could be saved. This workflow produced 3,000 complete annotation records and 15,000 individual distortion judgments overall. Annotators viewed only the distorted image: they were not shown the paired reference image or the decisions of the other annotators, and no consensus meeting was subsequently used to replace disagreements with a single agreed label. Rather than forcing the three ratings into a single hard ground-truth class, all individual annotations were retained during model evaluation. This is particularly important for visually ambiguous cases, where disagreement between annotators represents genuine uncertainty in the image rather than necessarily indicating annotation error. The complete rubric, interface, and presentation conditions are reported in the Supplementary Material \ref{supp:annotation_protocol}.

\subsection{Annotation Agreement and Assessment Metrics} 
\label{subsec:annotation_metrics} 

The annotation analysis was designed to answer two separate questions. First, do the annotators agree on whether a distortion is present at all? Second, when they agree that it is present, do they also agree on its severity? Separating these questions is important because distinguishing \textit{none} from \textit{present} is generally easier than distinguishing, for example, \textit{low} from \textit{medium} severity. 

\subsubsection{Distortion Presence Agreement} 

For presence analysis, the four severity categories were converted into a binary representation. The label \textit{none} was treated as distortion absent, whereas \textit{low}, \textit{medium}, and \textit{high} were grouped as distortion present. With three annotators, an image can therefore show either unanimous agreement (3--0) or a split regarding distortion presence (2--1). We first report the observed pairwise agreement, defined as the average percentage of identical binary decisions across the three possible annotator pairs. 




\subsubsection{Severity Agreement} 
The second analysis retains the complete four-level scale (\textit{none}, \textit{low}, \textit{medium}, and \textit{high}). For each image and distortion, the three human ratings are classified as unanimous when all annotators select the same severity, majority when two annotators select the same severity, and no majority when all three select different grades. Exact four-class agreement measures how often two annotators select precisely the same severity. However, it treats all disagreements equally. For example, predicting \textit{low} instead of \textit{medium} is counted in the same way as predicting \textit{none} instead of \textit{high}. But since the labels have a natural order, we additionally consider quadratic-weighted kappa (QWK), a metric that penalizes larger disagreements more strongly \cite{cohen1968weighted}: 

\begin{equation} 
    \kappa_w = 1- \frac{\sum_{a,b}w_{ab}O_{ab}} {\sum_{a,b}w_{ab}E_{ab}}, 
    \label{eq:qwk} 
\end{equation}

where $O_{ab}$ is the observed frequency of ratings $a$ and $b$, $E_{ab}$ is the corresponding frequency expected by chance, and 

\begin{equation} 
w_{ab}=\frac{(a-b)^2}{(K-1)^2} 
\end{equation}

defines the disagreement penalty for $K=4$ ordered severity levels. Thus, adjacent disagreements receive a smaller penalty than errors spanning several severity levels. Conditional QWK is also reported after restricting the analysis to cases where both annotators agree that the distortion is present. This isolates disagreement about severity from disagreement about presence. 

\subsubsection{MLLM Assessment Metrics} 

The MLLM predictions are evaluated directly against all three human annotations rather than against a majority-vote label. This strategy allows ambiguous images to contribute proportionally to the evaluation instead of imposing an artificial single ground truth. Distortion detection is first evaluated using presence balanced accuracy. Model predictions are converted to the same binary representation used for the annotators, with \textit{none} representing absence and all other severity levels representing presence. Balanced accuracy is calculated as the mean of sensitivity and specificity. Balanced accuracy is preferred to standard accuracy because the frequency of present and absent distortions differs substantially across distortion types. To evaluate the complete ordinal severity prediction, we use the ranked probability score (RPS) \cite{gneiting2007scoring}. In the present benchmark, each model returns one discrete severity level. Under this setting, RPS simplifies to the normalized distance between the model prediction $\hat{y}$ and a human rating $y$: 

\begin{equation} 
\mathrm{RPS}=\frac{|\hat{y}-y|}{K-1}, \qquad K=4, 
\label{eq:rps_simple} 
\end{equation} 

where the four severity levels are encoded as $0$ (\textit{none}), $1$ (\textit{low}), $2$ (\textit{medium}), and $3$ (\textit{high}). A correct prediction therefore gives $\mathrm{RPS}=0$, an adjacent error gives $\mathrm{RPS}=1/3$, and the largest possible error, such as predicting \textit{high} when the annotator selected \textit{none}, gives $\mathrm{RPS}=1$. For interpretability, results are reported as $1-\mathrm{RPS}$ so that larger values always indicate better performance. The joint ordinal score averages this quantity over all images and all three annotators: 

\begin{equation}
\begin{aligned}
S_{\mathrm{joint}}
&= 1- \frac{1}{(K-1)NR}
\sum_{i=1}^{N} \sum_{r=1}^{R}
|\hat{y}_i-y_{ir}|, \\
&\qquad K=4,\quad R=3.
\end{aligned}
\label{eq:joint_ordinal_score}
\end{equation}

where $N$ is the number of evaluated images and $R$ is the number of annotators. A score of 1 therefore indicates perfect agreement with all annotators, while progressively larger severity errors reduce this score. 

A second ordinal metric, conditional severity $1-\mathrm{RPS}$, evaluates only cases in which both the model and the corresponding annotator consider the distortion present. This metric answers a different question: once the model has detected a distortion, how accurately does it estimate whether its severity is low, medium, or high? Because a model that rarely predicts any distortion could otherwise obtain a misleading conditional score from only a few detections, conditional severity is always reported together with detection coverage, defined as the fraction of relevant cases for which the model predicts a non-\textit{none} severity. 

All metrics are first calculated separately for each of the five distortion types and then macro-averaged so that every distortion contributes equally to the final score. To estimate the sensitivity of these results, we use 1,000 image-level bootstrap replicates \cite{efron1994bootstrap}. In each replicate, images are randomly resampled with replacement from the benchmark and the evaluation metrics are recomputed. The resulting distribution is used to calculate the reported 95\% confidence intervals. The same resampled image indices are used for every model and prompt within each replicate, ensuring that model and prompt comparisons are made on identical bootstrap samples.

\subsubsection{Free-Text Evidence Analysis}
\label{subsubsec:evidence_analysis_method}

In addition to the structured severity and confidence outputs, each distortion assessment includes a short free-text evidence description of the visible image features supporting the prediction. This field is not used to compute the numerical assessment metrics or to guide the human annotations; instead, it provides an interpretable record of the visual cues associated with each predicted severity level. We analyze these descriptions separately to determine whether their linguistic characteristics provide additional information about model behavior. In particular, we focus on hedging expressions such as ``slight'', ``subtle'', ``faint'', or ``possibly'', which are commonly associated with linguistic uncertainty. However, such verbal cues should not automatically be interpreted as calibrated measures of model confidence \cite{yona2024uncertainty}.

\begin{table*}[!t]
\centering
\caption{Unified models and their observed inference cost, averaged over the four prompt variants of each stage (P0--P3 for assessment and R0--R3 for restoration). Dates below the model names indicate the Hugging Face release month of the exact checkpoint used in the benchmark. Parameter counts are obtained from local checkpoint tensor headers. Assessment token counts represent the mean total processor tokens per call, while restoration token counts represent text-conditioning tokens only and exclude visual and latent image representations. Latency is the mean elapsed inference time per completed call, peak GPU memory is the maximum allocated device memory, and Runtime is the mean total inference time per prompt sweep. For S1-Omni-Image, peak GPU memory is summed across two NVIDIA A100 80~GB GPUs; the remaining unified MLLMs use one A100 80~GB GPU.}
\label{tab:models}

\footnotesize
\setlength{\tabcolsep}{2.5pt}

\begin{adjustbox}{max width=\textwidth}
\begin{tabular}{@{}lccc cccc cccc@{}}
\toprule
& & & &
\multicolumn{4}{c}{\textbf{Assessment}} &
\multicolumn{4}{c}{\textbf{Restoration}} \\
\cmidrule(lr){5-8}
\cmidrule(lr){9-12}

\textbf{Model} &
\textbf{Params} &
\textbf{Size (GB)} &
\textbf{Max side} &
\textbf{Tokens} &
\makecell{\textbf{Latency} \textbf{(s)}} &
\makecell{\textbf{Peak GPU}\\\textbf{memory (GiB)}} &
\textbf{Runtime} &
\textbf{Tokens} &
\makecell{\textbf{Latency} \textbf{(s)}} &
\makecell{\textbf{Peak GPU}\\\textbf{memory (GiB)}} &
\textbf{Runtime} \\
\midrule

\makecell[l]{BAGEL \cite{deng2025bagel} \\\scriptsize Jan. 2026}
& 14.69B & 29.56 & 1024 px
& 4227 & 9.91 & 27.97 & 2h 46m
& 444 & 44.00 & 30.52 & 1h 14m \\

\makecell[l]{InternVL-U \cite{tian2026internvlu} \\\scriptsize Mar. 2026}
& 4.28B & 8.79 & 1024 px
& 1284 & 15.12 & 8.20 & 4h 13m
& 533 & 10.60 & 14.31 & 0h 20m \\

\makecell[l]{MammothModa2 \cite{shen2025mammothmoda2} \\\scriptsize Dec. 2025}
& 15.11B & 51.06 & 512 px
& 1476 & 11.70 & 61.15 & 3h 16m
& 677 & 21.85 & 70.06 & 0h 37m \\

\makecell[l]{S1-Omni-Image \cite{li2026s1omniimage} \\\scriptsize Jun. 2026}
& 53.96B & 107.94 & 1024 px
& 1760 & 15.77 & 101.24 & 4h 24m
& 627 & 52.68 & 106.13 & 1h 29m \\

\makecell[l]{SenseNova-U1 \cite{sensenova2026u1} \\\scriptsize May 2026}
& 17.55B & 35.22 & 1024 px
& 1436 & 20.56 & 33.80 & 5h 43m
& 806 & 18.66 & 34.70 & 0h 32m \\

\bottomrule
\end{tabular}
\end{adjustbox}

\end{table*}

An initial lexicon of 19 candidate hedge expressions was defined before analysis. To avoid conclusions based on rarely occurring terms, detailed analysis was restricted to expressions appearing at least 400 times in the evidence corpus. Six terms satisfied this criterion: ``slight'', ``subtle'', ``faint'', ``appear'', ``slightly'', and ``mild''. We first compare their frequency between correct and incorrect severity predictions and then examine whether hedge use is associated more strongly with prediction correctness or with the severity level produced by the model.

Because these variables are categorical, the associations are evaluated using chi-square tests of independence. Statistical significance is reported using the corresponding $p$-value, while association strength is quantified using Cram\'er's $V$, which ranges from 0 for no association to 1 for a perfect association. Two comparisons are performed: hedge presence versus prediction correctness (correct/incorrect), and hedge presence versus predicted severity (\textit{none}, \textit{low}, \textit{medium}, \textit{high}). This allows us to determine whether hedging is more closely related to prediction accuracy or simply to the severity level being described.

\subsection{Unified MLLMs}
\label{subsec:unified_mllms}

Five open-weight unified multimodal large language models (MLLMs) were selected for evaluation. Unlike conventional vision-language models that only interpret images, the selected frameworks support both image understanding and image generation or editing within the same model family. This property is essential for the proposed benchmark, since the same model is required first to assess the water-induced distortions and then to generate a restored version of the image.

BAGEL uses a Mixture-of-Transformer-Experts (MoT) architecture combining a Qwen2.5 language backbone with separate pixel-level and semantic-level visual encoders under a shared next-group-token-prediction objective \cite{deng2025bagel}. InternVL-U combines an MLLM with a modality-specific MMDiT generation head in a unified-context architecture \cite{tian2026internvlu}. MammothModa2 integrates a Qwen3-VL understanding backbone, autoregressive generation experts, and a dense diffusion transformer \cite{shen2025mammothmoda2}. S1-Omni-Image conditions image generation on hidden states produced by a scientific vision-language reasoning backbone using a think-before-generate strategy \cite{li2026s1omniimage}. SenseNova-U1 adopts the NEO-unify Mixture-of-Transformers architecture for native multimodal understanding and pixel generation \cite{sensenova2026u1}.

In all cases, assessment and restoration are performed within the same locally hosted framework. All publicly released checkpoints are evaluated with their pretrained weights kept frozen throughout the experiments. No fine-tuning, parameter-efficient adaptation, or task-specific training is performed. The exact checkpoint revision used for each model was fixed throughout the benchmark. Table~\ref{tab:models} summarizes the main model characteristics together with the observed computational requirements and the release date of each checkpoint.

\subsection{Prompt Engineering}
\label{subsec:prompt_design}

Prompt design was treated as a controlled experimental factor rather than as an implementation detail. Previous prompt-engineering studies have shown that task instructions, contextual roles, explicit definitions, and task decomposition can substantially modify model behavior without changing the underlying model parameters \cite{sahoo2024prompt}. Similarly, structured and stepwise prompting can improve reasoning on multi-stage tasks \cite{wei2022chain,kojima2022large}. We therefore designed four progressively structured prompts for assessment (P0--P3) and four corresponding prompts for restoration (R0--R3), see Table \ref{tab:prompt_design}.

All prompts follow the same zero-shot and reference-free protocol. Models receive only the distorted input image and the corresponding text instruction. They are never shown the paired reference image, a worked example, an in-context demonstration, or a description of the true seabed. Consequently, the prompt variants differ only in how the task is formulated and not in the amount of information made available to the model.

For assessment, P0 provides a minimal direct instruction, P1 introduces a domain-expert role, P2 adds explicit physical definitions and severity criteria, and P3 decomposes the assessment into a structured internal checklist. All four prompts require the same five distortions to be evaluated and return the same predefined JSON structure. For restoration, R0 directly requests correction according to the model-generated diagnosis, R1 introduces an underwater-optics expert role, R2 explicitly asks the model to formulate a restoration plan before generating the image, and R3 emphasizes minimum intervention and preservation of the original scene. The complete operative prompts are reported in Table~\ref{tab:prompt_design}.

\begin{table*}[!t]
\centering
\caption{Assessment and restoration prompt design. The fixed assessment JSON schema and the restoration diagnosis and priority placeholders are applied consistently to all prompt variants within their corresponding stage. [ID], [ORDERED ACTIONS], [PRIORITY PLAN] and [DIAGNOSIS] are retrieved from the assessment output.}
\label{tab:prompt_design}
\scriptsize
\setlength{\tabcolsep}{3pt}
\renewcommand{\arraystretch}{1.12}
\begin{tabular}{@{}llp{0.14\textwidth}p{0.20\textwidth}p{0.43\textwidth}@{}}
\toprule
\textbf{Task} & \textbf{Name} & \textbf{Method} & \textbf{Description} & \textbf{Operative prompt} \\
\midrule
Assessment & P0 Direct & Zero-shot baseline & Minimal instruction; establishes performance without domain scaffolding. & ``Analyze this UAV shallow-water image. For caustics, refraction, sun glint, turbidity, and light absorption, assign none, low, medium, or high. Select the dominant distortion and overall restoration need. Judge only visible evidence; return one valid JSON object.'' \\
Assessment & P1 Expert & Role/context & Adds a shallow-water remote-sensing and aquatic-optics role without definitions. & ``Assess the water-induced degradation in this UAV shallow-water image as a remote-sensing scientist specializing in aquatic optics. Grade the five distortions from visible evidence and identify what most limits seabed interpretability. Return one valid JSON object.'' \\
Assessment & P2 Rubric & Definition-guided & Adds physical definitions, severity anchors, and counterexamples for common confounders. & ``Assess using operational definitions: caustics are focused bright networks; refraction is water-surface geometric displacement; sun glint is saturated specular reflection; turbidity is a scattering veil; light absorption is broad attenuation/color loss. Use none/low/medium/high anchors and do not confuse seabed appearance with an artifact. Return one valid JSON object.'' \\
Assessment & P3 Checklist & Task decomposition & Orders radiometric, illumination, geometry, and visibility checks; only the final answer is exposed. & ``Internally check: A) global radiometry, B) caustics versus sun glint, C) geometric instability, and D) scattering/visibility. Then grade all five distortions and restoration need. Do not output chain-of-thought; return only the final JSON.'' \\
\midrule
Restoration & R0 Priority-direct & Direct editing & Applies the score-derived correction order with the least additional scaffolding. & ``Restore image [ID] into a plausible undistorted view of the same scene. Apply [ORDERED ACTIONS] from highest to lowest priority using [DIAGNOSIS] and [PRIORITY PLAN]. Correct only diagnosed artifacts and return one natural restored image.'' \\
Restoration & R1 Expert optics & Role/context & Requests physically conservative, artifact-specific corrections. & ``As an underwater-optics scientist, create a scientifically plausible undistorted version. Correct higher-scoring distortions first: suppress glint without inventing seabed; stabilize refraction; flatten caustics; reduce turbidity; then correct absorption. Preserve radiometric and geometric realism.'' \\
Restoration & R2 Reasoning-guided & Plan then edit & Requests an evidence-grounded restoration plan before synthesis. & ``Reason through restoration before editing: verify each diagnosed artifact, follow [PRIORITY PLAN], choose strength consistent with severity/confidence, protect real seabed content, then generate the final image. Do not invent occluded detail or alter scene layout.'' \\
Restoration & R3 Conservative & Minimum-change editing & Applies the strongest scene-preservation constraint. & ``Generate the closest possible counterfactual of the identical scene and viewpoint without the diagnosed distortions. Apply only [ORDERED ACTIONS]. Keep unaffected regions unchanged and reject generic grading, synthetic texture, halos, smoothing, or unsupported reconstruction.'' \\
\bottomrule
\end{tabular}
\end{table*}

All four assessment prompts were applied once to each of the 1,000 images for every model, resulting in 20,000 model responses and 100,000 individual distortion predictions across the complete five-model benchmark. Accordingly, comparisons among P0--P3 over all 1,000 images are interpreted descriptively as prompt-sensitivity analyses rather than as fully held-out prompt-selection estimates.

\subsection{Diagnosis-to-Restoration Prioritization}
\label{subsec:restoration_prioritization}

The restoration stage uses the model's own assessment output rather than human annotations. This diagnosis-guided formulation follows recent MLLM-based image-restoration agents, where degradation perception is used to identify the required restoration operations and determine their execution order \cite{chen2024restoreagent,zhou2025qagent}. In contrast to these agent-based approaches, which typically use the MLLM to select or orchestrate external restoration tools, our framework retains both diagnosis and image generation within the same unified MLLM family. For each image, the selected assessment prompt produces a severity and a self-reported confidence value for each of the five distortion types. These predictions are then converted into an ordered restoration plan that explicitly indicates which distortions should be addressed first.

Severity is mapped onto an ordinal numerical scale,
\begin{equation}
S^{(d)} \in \{0,1,2,3\},
\end{equation}
corresponding respectively to \textit{none}, \textit{low}, \textit{medium}, and \textit{high}. Each distortion is also assigned a fixed base priority $P^{(d)}$, representing the preferred order of correction rather than the physical magnitude of the distortion. Finally, the model's self-reported confidence $C^{(d)}$ ranges from 0 to 1.
For each distortion $d$, these three quantities are combined using the following heuristic prioritization rule:
\begin{equation}
S_c^{(d)}=P^{(d)}S^{(d)}C^{(d)}.
\label{eq:correction_score}
\end{equation}

Equation~\eqref{eq:correction_score} is an heuristic used only to obtain a consistent ordering of the diagnosed corrections; it is not intended as a calibrated physical model of distortion importance. The priority values were fixed before restoration evaluation and were not optimized against restoration performance. Their ordering reflects the intended correction sequence: sun glint receives the highest priority because strong specular reflection can obscure or saturate seabed information; refraction follows because it alters apparent scene geometry; caustics and turbidity receive equal intermediate priority because they primarily affect illumination and visibility; and light absorption is assigned the lowest active priority because it represents a broader radiometric attenuation that can be addressed after more localized or geometrically disruptive effects. The numerical values $5$, $4$, $3$, $3$, and $2$ therefore encode this operational preference rather than measured physical magnitudes.

Under this heuristic, a distortion receives a higher correction score when it has a higher predefined priority, is assigned a greater severity level, and is accompanied by a larger self-reported confidence value. Conversely, a distortion predicted as \textit{none} receives a severity score of zero and is excluded from the restoration plan. The active distortions are then sorted in decreasing order of $S_c^{(d)}$ and inserted into the restoration prompt.

\begin{table}[!t]
\centering
\caption{Ordinal severity scores and fixed restoration priorities used to construct the diagnosis-guided restoration plan. The priority values determine correction order and should not be interpreted as measures of physical distortion magnitude.}
\label{tab:distortion_priority_scoring}
\small
\begin{tabular}{lc@{\qquad}lc}
\toprule
\textbf{Severity level} & \textbf{$S^{(d)}$} & \textbf{Distortion type} & \textbf{$P^{(d)}$} \\
\midrule
None & 0 & Sun glint & 5 \\
Low & 1 & Refraction & 4 \\
Medium & 2 & Caustics & 3 \\
High & 3 & Turbidity & 3 \\
-- & -- & Light absorption & 2 \\
-- & -- & None & 0 \\
\bottomrule
\end{tabular}
\end{table}

For example, a \textit{medium} sun-glint prediction with confidence $0.9$ produces $S_c=5\times2\times0.9=9.0$, whereas a \textit{high} turbidity prediction with confidence $0.8$ produces $S_c=3\times3\times0.8=7.2$. Sun glint would therefore be addressed first despite its lower predicted severity. This mechanism does not determine how the image must be modified; it only provides the unified MLLM with a consistent ordering of the diagnosed corrections.

For a given model and image, all four restoration prompts R0--R3 receive the same model-generated diagnosis and the same ordered restoration plan. No human annotation or reference-derived information is introduced at this stage. The prompt variants therefore test different ways of instructing the generation component while holding the underlying diagnosis constant. The same latent seed is also retained across R0--R3 for a given model--image pair wherever supported by the corresponding framework.

\subsection{Experimental Setup}
\label{subsec:experimental_setup}

All experiments were conducted strictly zero-shot using publicly released pretrained checkpoints. Model parameters remained frozen throughout the benchmark, with no fine-tuning, gradient updates, LoRA adaptation, or in-context examples. During both assessment and restoration, each model received only the distorted image and the corresponding text instruction. Paired reference images were withheld throughout inference and used only afterward for quantitative evaluation.

For assessment, each model receives one RGB image at either $512\!\times\!512$ pixels for Sea-Undistort or $1000\!\times\!1000$ pixels for KAUSTics, together with one of the P0--P3 prompts. All runs use batch size one with no quantization. Text generation uses greedy decoding with temperature 0, top-$p=1$, with a maximum of 700 tokens. The required output reports, for each of the five distortions, the predicted severity, a self-reported confidence value, and a short visible-evidence description, together with the dominant distortion and the overall restoration requirement. 

For restoration, the same distorted image is combined with the model-generated diagnosis and the ordered correction plan described in Section~\ref{subsec:restoration_prioritization}. The output is one restored RGB image using the R0--R3 prompts. Output dimensions follow the input image subject to the maximum image size supported by each model, as reported in Table~\ref{tab:models}. InternVL-U and MammothModa2 use 20 diffusion steps, while BAGEL, S1-Omni-Image, and SenseNova-U1 use 50. Restoration experiments used 100 images per model. The restoration protocol specifies the same 100-image subset for all seven models, comprising 50 KAUSTics and 50 Sea-Undistort images. The subset was selected to preserve the per-distortion severity distribution while requiring at least one active distortion.

To complement the unified MLLMs, we evaluate two restoration-only baselines: InstructIR~\cite{conde2024instructir}, which conditions a convolutional restoration network on a compact text encoder, and PromptFix~\cite{yu2024promptfix}, an instruction-conditioned latent diffusion model. Their public checkpoints are used without adaptation to this benchmark. Both receive the existing InternVL-U P3 assessment and ordered correction plan from assessment. The two baselines use the same four R0--R3 intents. They are therefore comparisons of the restoration stage under fixed upstream conditioning. Their original training may use paired supervision; zero-shot here refers to application to our benchmark with frozen weights and withheld reference images, rather than unsupervised model training.


As a fixed-seed repeatability check, all assessment and restoration experiments were repeated three times using identical hyperparameters, seeds, decoding settings, and generation settings. The assessment responses and restored outputs were identical across all three repetitions, confirming deterministic repeatability under the fixed experimental configuration. Because the three repetitions were identical, the reported results correspond to one representative repetition. Hardware requirements and measured computational costs are summarized in Table~\ref{tab:models}.

\subsection{Restoration Metrics}

Restoration quality is evaluated against the paired reference images using peak signal-to-noise ratio (PSNR), structural similarity index (SSIM), multi-scale structural similarity index (MS-SSIM), and learned perceptual image patch similarity (LPIPS) \cite{wang2004ssim,wang2003msssim,zhang2018lpips}. Higher PSNR, SSIM, and MS-SSIM indicate greater similarity to the reference, whereas lower LPIPS indicates better perceptual similarity. Outputs with differing dimensions are resized to the paired reference grid using Lanczos resampling. Each restored image is compared with both its paired reference and the distorted-input baseline, so restoration performance is primarily reported as the change relative to leaving the input unmodified. Confidence intervals for these changes are estimated using 1,000 image-level bootstrap resamples, with the same resampled image indices used across prompts. The 2.5th and 97.5th percentiles of the resulting distributions are reported as 95\% confidence intervals.

\subsubsection{Feature and Segmentation Diagnostics}
\label{subsec:downstream_method}

For each model, we select its own three distinct images per dataset with the largest SSIM gain over the input, first taking the best R0--R3 output per image. Ties are resolved by image identifier and prompt identifier. These 42 evaluation-selected restorations constitute a best-case diagnostic. The corresponding input and reference are processed identically. This diagnostic uses a common grid within each triplet, independently of the original full-reference metric preprocessing.

We count SIFT keypoints~\cite{lowe2004sift} using the library's default settings and masks generated by the frozen Segment Anything ViT-B model~\cite{kirillov2023segany}, using a $32\!\times\!32$ point grid, predicted-IoU threshold 0.88, stability threshold 0.95, and no crop pyramid or model adaptation. Input and restored counts are compared with the corresponding reference counts. Larger keypoint or mask counts alone need not imply better restoration: texture artifacts can create keypoints and oversegmentation can increase mask counts. These diagnostics therefore provide simple measures of reference-like structural content rather than direct evidence of downstream task performance.

\subsection{Dataset-Stratified Analysis}
\label{subsec:dataset_stratified}

Assessment and restoration metrics are additionally computed separately for KAUSTics and Sea-Undistort from the same outputs. Assessment retains the evaluated prompts and all three human labels, with macro-averaging over distortion types. A per-distortion metric is undefined when its required reference class or detected-positive subset is absent; undefined values are excluded from macro-averages. Confidence intervals use 1,000 image-bootstrap resamples within each dataset, preserving the three annotators and five distortions together and sharing image draws across model--prompt configurations. For restoration, the metric-specific prompt selected from each pooled 100-image cohort is held fixed for the dataset-specific table. This separates dataset effects without further prompt selection. Dataset differences jointly reflect image formation, scene content, native resolution, and distortion prevalence.

\section{Results}
\label{sec:results}

The results follow the benchmark stages: human annotation reliability, model assessment and evidence, and restoration performance, including the two restoration-only baselines.

\subsection{Human Annotation Reliability}
\label{subsec:annotation_analysis_results}

\begin{figure*}[!t]
\centering
\includegraphics[width=0.98\textwidth]{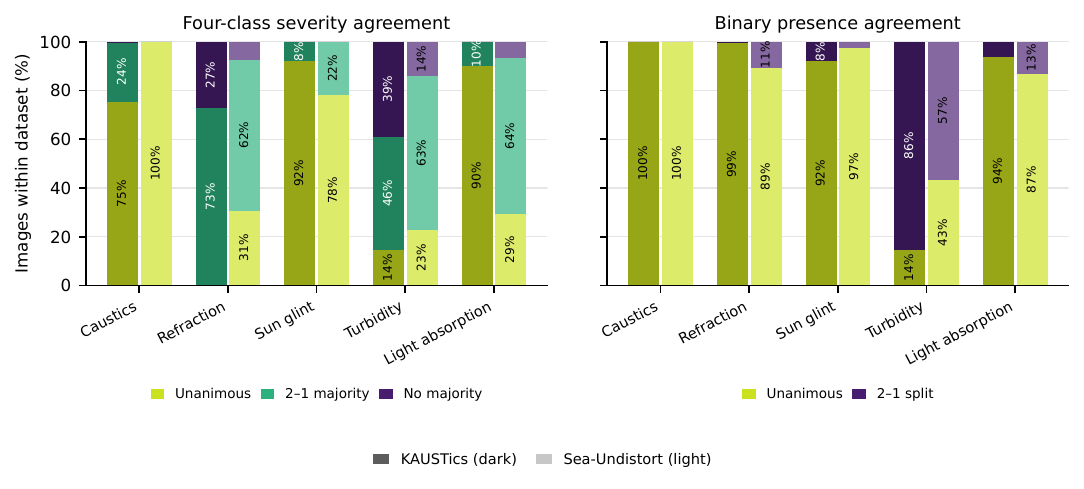}
\caption{Agreement among the three annotators for each water-induced distortion. Left: four-class severity agreement, categorized as unanimous, 2--1 majority, or no majority. Right: binary presence agreement after grouping \textit{low}, \textit{medium}, and \textit{high} as present. Each distortion has paired bars for KAUSTics (dark; 600 images) and Sea-Undistort (light; 400 images). Percentages are normalized within each dataset and labels are shown for segments of at least 8\%.}
\label{fig:annotator_agreement}
\end{figure*}

Before evaluating the MLLMs, it is necessary to establish how consistently the five water-induced distortions can be identified by human observers. The annotation results show that this difficulty is not uniform across distortion types. Table~\ref{tab:annotator_reliability} reports the corresponding inter-annotator reliability statistics, while Fig.~\ref{fig:annotator_agreement} shows image-level agreement by dataset.

\begin{table}[!h]
\centering
\caption{Inter-annotator reliability over the 1,000-image benchmark. QWK is conditional on both annotators judging the distortion present. Higher values indicate stronger agreement.}
\label{tab:annotator_reliability}
\small
\setlength{\tabcolsep}{3pt}
\begin{adjustbox}{max width=\columnwidth}
\begin{tabular}{@{}lccc@{}}
\toprule
\textbf{Distortion type} &
\textbf{Pres. agr. $\uparrow$} &
\textbf{4-class exact $\uparrow$} &
\textbf{QWK $\uparrow$} \\
\midrule
Caustics         & \textbf{1.000} & 0.901          & 0.157 \\
Refraction       & 0.969                   & 0.351          & 0.161 \\
Sun glint        & 0.961                  & \textbf{0.909} & 0.786 \\
Turbidity        & 0.507                  & 0.355          & 0.115 \\
Light absorption & 0.939                 & 0.762          & \textbf{0.819} \\
\bottomrule
\end{tabular}
\end{adjustbox}
\end{table}

Binary presence agreement was complete for caustics but substantially lower for turbidity (0.507), indicating that annotators frequently disagreed on whether turbidity was present at all. A straightforward explanation is that caustics and sun glint generally produce spatially distinctive visual patterns, whereas turbidity appears as a more gradual loss of contrast and visibility that can be difficult to separate from naturally low-contrast seabed appearance. Caustics unanimity reflects dataset-specific prevalence: all three annotators marked it present in every KAUSTics image and absent in every Sea-Undistort image. Turbidity remained ambiguous within both datasets, with unanimous binary agreement of 14.5\% in KAUSTics and 43.2\% in Sea-Undistort.


Recognizing presence did not ensure agreement on severity. Refraction had binary agreement of 0.969 but exact four-class agreement of only 0.351. Thus, annotators generally recognized that refraction was present but often assigned different severity levels. For caustics, unanimous absence labels in Sea-Undistort increase pooled exact agreement (0.901), whereas conditional QWK (0.157) concerns severity agreement beyond chance only among present cases, where ratings are concentrated in the high category. Conditional QWK, which evaluates severity agreement only when both annotators judged the distortion present, was highest for light absorption (0.819) and sun glint (0.786). These distortion-dependent differences support retaining all three labels rather than imposing consensus.

\subsection{MLLM Assessment Performance}
\label{subsec:assessment_results}

Assessment was evaluated for all five unified MLLMs and four prompts over the complete 1,000-image benchmark.

\subsubsection{Prompt Sensitivity}
\label{subsec:assessment_per_prompt}

\begin{figure*}[!t]
\centering
\includegraphics[width=0.9\textwidth]{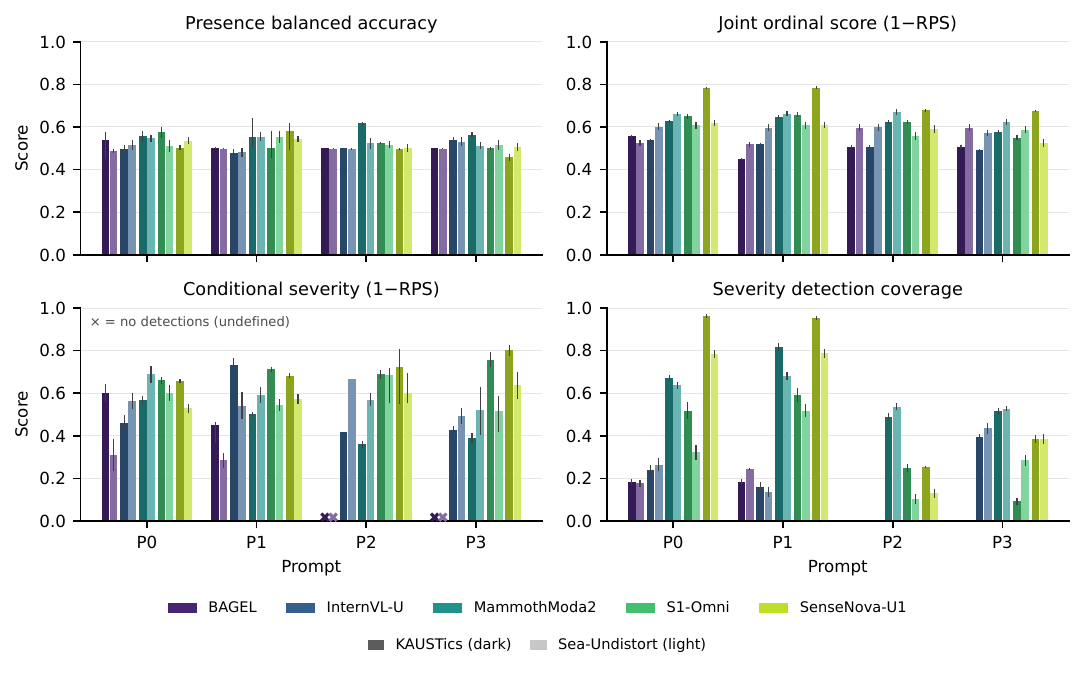}
\caption{Assessment performance across the four prompt variants, separated by dataset. Each model--prompt pair uses dark KAUSTics and light Sea-Undistort bars. Scores are macro-averaged over defined distortion-specific metrics; caustics presence BA is undefined within both datasets because one reference class is absent. Error bars show within-dataset image-bootstrap 95\% confidence intervals. Higher values indicate better performance for all metrics. An `$\times$' in the conditional-severity panel indicates zero detection coverage, for which conditional severity is undefined.}
\label{fig:assessment_prompt_metrics}
\end{figure*}

The first question is whether progressively adding role information, physical definitions, or task decomposition improves distortion assessment. Figure~\ref{fig:assessment_prompt_metrics} summarizes the four principal assessment metrics for every prompt--model combination, and Table~\ref{tab:prompt_sensitivity_ranges} reports the corresponding best-to-worst variation across prompts.

\begin{table}[!h]
\centering
\caption{Prompt sensitivity over all 1,000 images. Ranges correspond to the difference between the best and worst prompt for each metric; parentheses identify the best-performing prompt.}
\label{tab:prompt_sensitivity_ranges}
\footnotesize
\begin{adjustbox}{max width=\columnwidth}
\begin{tabular}{@{}lcc@{}}
\toprule
\textbf{Model} & \textbf{Presence BA range} & \textbf{Joint range} \\
\midrule
BAGEL \cite{deng2025bagel} & 0.037 (P0) & 0.067 (P0) \\
InternVL-U \cite{tian2026internvlu} & 0.043 (P3) & 0.040 (P0) \\
MammothModa2 \cite{shen2025mammothmoda2} & 0.052 (P1) & 0.057 (P1) \\
S1-Omni-Image \cite{li2026s1omniimage} & 0.052 (P1) & 0.073 (P1) \\
SenseNova-U1 \cite{sensenova2026u1}& \textbf{0.062} (P1) & \textbf{0.103} (P0) \\
\bottomrule
\end{tabular}
\end{adjustbox}
\end{table}

Changing the prompt altered presence balanced accuracy by up to 0.062 and the joint ordinal score by 0.103. SenseNova-U1 showed the largest variation, performing best under P1 for presence BA but P0 for the joint score. MammothModa2 and S1-Omni-Image also favored P1 for presence BA, whereas InternVL-U favored P3. BAGEL's detection coverage fell to zero under P2 and P3, making conditional severity undefined. Thus, additional prompt structure does not produce a consistent improvement across models and can instead alter the underlying prediction behavior.

\begin{figure*}[!t]
\centering
\includegraphics[width=\textwidth]{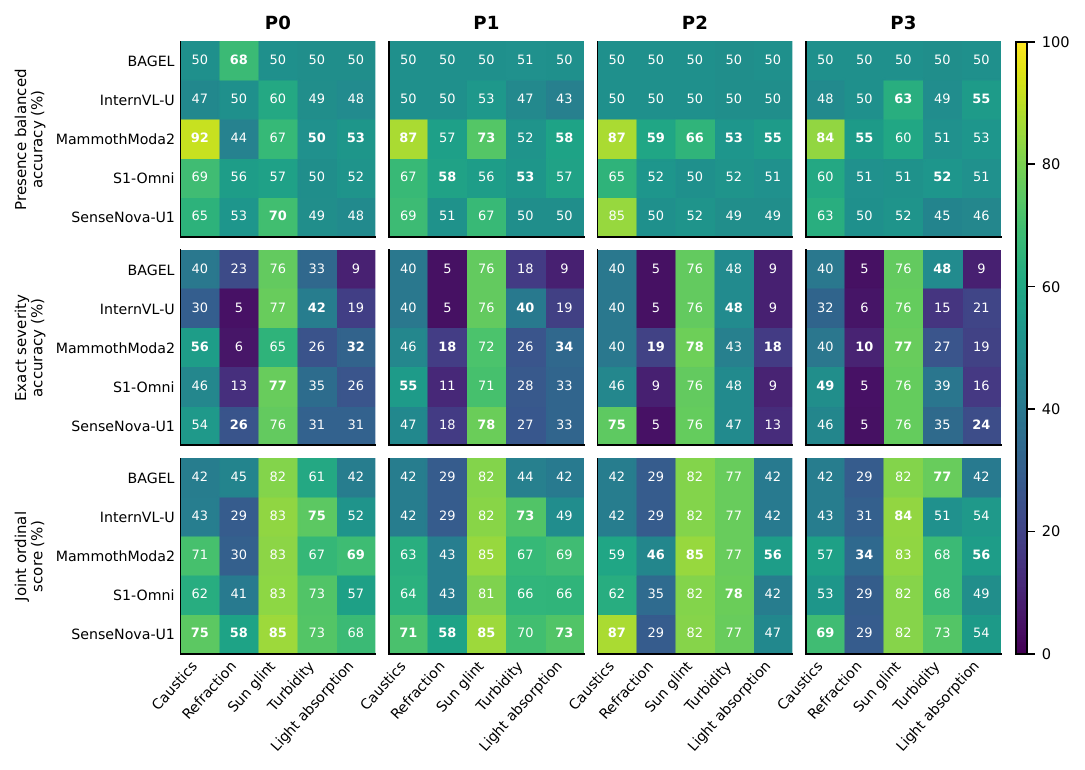}
\caption{Assessment performance by prompt, model, and distortion over the pooled 1,000 images. Rows show presence balanced accuracy, exact four-class severity accuracy, and the joint ordinal score ($1-\mathrm{RPS}$). Values are reported as percentages. Both presence classes occur in the pooled data for every distortion, so all presence BA values are defined. Bold numbers identify the highest score across models for each distortion within each metric--prompt panel.}
\label{fig:assessment_prompt_heatmap}
\end{figure*}

Distortion-specific results show a similar contrast (Fig.~\ref{fig:assessment_prompt_heatmap}). Sun-glint presence BA ranged from approximately 50--73\%, refraction from 44--68\%, and turbidity from 45--53\%. Several configurations therefore provided little discrimination between presence and absence despite class imbalance. Turbidity's joint ordinal score nevertheless reached 78\%, illustrating why exact severity accuracy and $1-\mathrm{RPS}$ need not vary together: adjacent severity errors receive only a partial penalty under the ordinal metric.

\subsubsection{Model-Level Assessment Performance}
\label{subsec:assessment_overall}

The prompt-level analysis shows that no single prompting strategy provides the strongest performance for every model and metric. To provide a compact model-level comparison, Table~\ref{tab:assessment_results} reports all four metrics at each model's best pooled presence-BA prompt: BAGEL P0, InternVL-U P3, and P1 for MammothModa2, S1-Omni-Image, and SenseNova-U1. In the pooled analysis, MammothModa2 had the highest presence BA (0.654), while SenseNova-U1 led the joint ordinal score (0.714) and detection coverage (0.837). S1-Omni-Image achieved the highest conditional severity score (0.655), but at lower coverage (0.520). BAGEL and InternVL-U had lower joint scores (0.543 and 0.523) and coverage (0.149 and 0.378), illustrating why conditional severity should be interpreted alongside detection coverage.

\begin{table*}[!t]
\centering
\caption{Assessment performance at each model's best pooled presence-BA prompt, separated into KAUSTics (K) and Sea-Undistort (S); image counts are in parentheses. The same prompt is used for both datasets and all metrics. Brackets give image-bootstrap 95\% confidence intervals conditional on that prompt. Bold marks the best estimate within each dataset.}
\label{tab:assessment_results}
\footnotesize
\setlength{\tabcolsep}{3pt}
\begin{adjustbox}{max width=\textwidth}
\begin{tabular}{llcccc}
\toprule
\textbf{Model (prompt)} & \textbf{Dataset} & \textbf{Presence BA $\uparrow$} & \textbf{Joint $1-$RPS $\uparrow$} & \textbf{Conditional $1-$RPS $\uparrow$} & \textbf{Coverage $\uparrow$} \\
\midrule
BAGEL \cite{deng2025bagel} (P0) & K (600) & 0.540 [0.450, 0.576] & 0.556 [0.548, 0.563] & 0.603 [0.560, 0.644] & 0.184 [0.170, 0.197] \\
 & S (400) & 0.489 [0.477, 0.499] & 0.525 [0.512, 0.538] & 0.308 [0.234, 0.385] & 0.177 [0.161, 0.193] \\
\addlinespace[2pt]
InternVL-U \cite{tian2026internvlu} (P3) & K (600) & 0.537 [0.522, 0.554] & 0.490 [0.483, 0.496] & 0.427 [0.405, 0.445] & 0.394 [0.379, 0.409] \\
 & S (400) & 0.531 [0.510, 0.552] & 0.573 [0.559, 0.587] & 0.492 [0.453, 0.531] & 0.436 [0.410, 0.461] \\
\addlinespace[2pt]
MammothModa2 \cite{shen2025mammothmoda2} (P1) & K (600) & 0.552 [0.504, 0.644] & 0.648 [0.639, 0.656] & 0.501 [0.488, 0.513] & 0.817 [0.800, 0.834] \\
 & S (400) & \textbf{0.552} [0.534, 0.574] & \textbf{0.663} [0.651, 0.674] & \textbf{0.590} [0.554, 0.630] & 0.681 [0.664, 0.699] \\
\addlinespace[2pt]
S1-Omni-Image \cite{li2026s1omniimage} (P1) & K (600) & 0.503 [0.455, 0.580] & 0.658 [0.646, 0.671] & \textbf{0.712} [0.700, 0.724] & 0.590 [0.557, 0.627] \\
 & S (400) & 0.551 [0.524, 0.579] & 0.607 [0.592, 0.621] & 0.546 [0.518, 0.574] & 0.517 [0.490, 0.547] \\
\addlinespace[2pt]
SenseNova-U1 \cite{sensenova2026u1} (P1) & K (600) & \textbf{0.580} [0.493, 0.620] & \textbf{0.785} [0.779, 0.791] & 0.683 [0.674, 0.693] & \textbf{0.952} [0.944, 0.960] \\
 & S (400) & 0.541 [0.527, 0.557] & 0.608 [0.595, 0.621] & 0.572 [0.548, 0.596] & \textbf{0.786} [0.766, 0.807] \\
\bottomrule
\end{tabular}
\end{adjustbox}
\end{table*}

These rankings varied by dataset. SenseNova-U1's joint score was 0.785 on KAUSTics and 0.608 on Sea-Undistort; InternVL-U scored 0.490 and 0.573. SenseNova-U1 led KAUSTics presence BA (0.580), whereas MammothModa2 was slightly higher than S1-Omni-Image on Sea-Undistort (0.552 versus 0.551). Caustics presence BA is undefined within each dataset because one reference class is absent. The resulting four-distortion BA averages are therefore not a weighted decomposition of the pooled five-distortion BA. Per-distortion severity distributions and confusion matrices are provided in Sections~\ref{app:severity_distributions} and \ref{app:confusion_matrices} of the Supplementary Material.

Pooled caustics presence BA may partly reflect recognition of the dataset rather than within-domain distortion detection. Excluding caustics while retaining the same prompts reduces MammothModa2's pooled BA from 0.654 to 0.600 (95\% CI $[0.577,0.621]$), still the highest estimate; the other models range from 0.543 to 0.560. Including caustics therefore raises the pooled estimate, but MammothModa2's advantage is not restricted to this distortion.

\subsubsection{Free-Text Evidence Analysis}
\label{subsec:assessment_reasoning}

\begin{figure}[!h]
\centering
\includegraphics[width=\columnwidth]{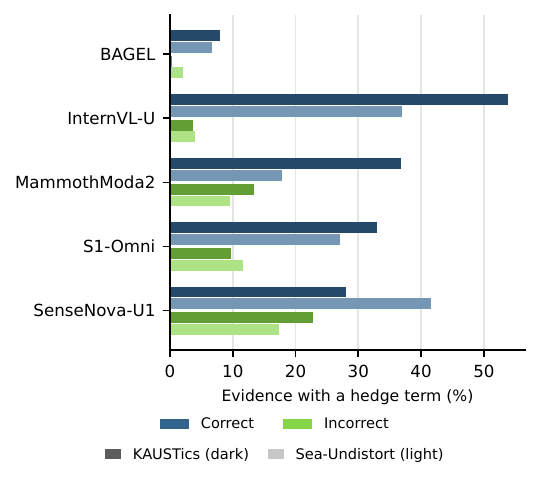}
\caption{Fraction of evidence descriptions containing at least one frequently used hedge term ('slight', 'subtle', 'faint', 'appear', 'slightly', 'mild'), separated by model, correctness, and dataset.}
\label{fig:hedging_by_model}
\end{figure}

The evidence descriptions allow linguistic expressions of uncertainty to be compared with prediction correctness. Of 19 candidate hedge expressions, six occurred frequently: ``slight'', ``subtle'', ``faint'', ``appear'', ``slightly'', and ``mild''. Hedge expressions were more frequent in correct than incorrect descriptions for all five models. InternVL-U showed the largest pooled difference: 47.4\% versus 3.7\%. Figure~\ref{fig:hedging_by_model} separates these rates by dataset; the pattern does not support treating hedge words as simple error indicators.

A straightforward explanation emerges from the semantics of the words themselves. Terms such as ``slight'', ``faint'', and ``mild'' directly describe weak visual effects and are therefore naturally associated with the \textit{low} severity level. Their occurrence may consequently reflect an appropriate description of subtle degradation rather than uncertainty about whether the degradation exists.

To determine what these words actually represent, we tested whether hedge use was more strongly associated with the predicted severity or with whether the prediction was correct. Both relationships were statistically significant ($p<0.001$), but their strengths differed substantially. Hedge use showed a strong association with predicted severity (Cram\'er's $V=0.58$), compared with a considerably weaker association with correctness ($V=0.24$). This indicates that those terms reflect the severity level assigned by the model.

\begin{figure}[!h]
\centering
\includegraphics[width=\columnwidth]{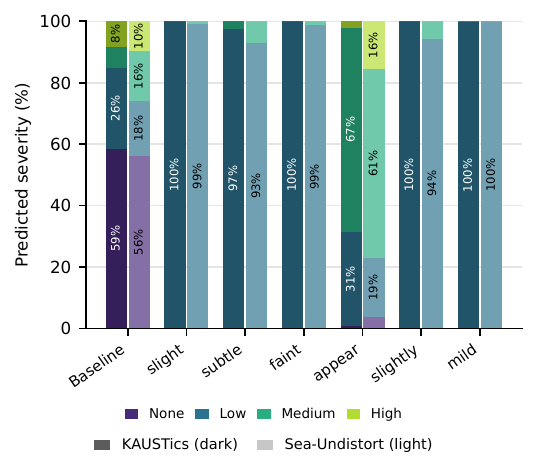}
\caption{Predicted severity distribution conditioned on each frequently occurring hedge term, with dark KAUSTics and light Sea-Undistort bars. The baseline uses all evidence descriptions within the corresponding dataset. Each bar sums to 100\%; segments of at least 8\% are labeled.}
\label{fig:hedge_severity_distribution}
\end{figure}

The severity distribution conditioned on individual hedge terms makes this relationship explicit. Five expressions occurred with \textit{low} predictions in 95.6--99.8\% of their pooled uses; ``appear'' instead occurred most often with \textit{medium} predictions (63.1\%). Figure~\ref{fig:hedge_severity_distribution} shows the dataset-specific distributions, which differed from the pooled baseline for every term. Further vocabulary and evidence-length analyses are in Section~\ref{app:evidence_length} of the Supplementary Material.

\subsection{Restoration Performance}
\label{subsec:restoration_results}

The assessment results establish whether the models can assess water-induced degradation, but the restoration stage of the benchmark investigates whether this recognition can be translated into restorations that reliably move the distorted observation closer to its paired reference. For quantitative remote sensing, restoration must preserve scene content while improving agreement with the reference. Restoration performance is therefore evaluated relative to each model's own distorted-input baseline. InstructIR and PromptFix additionally provide restoration-only comparisons using the fixed InternVL-U diagnosis and correction plan described in Section~\ref{subsec:experimental_setup}.

\subsubsection{Quantitative Restoration Performance}
\label{subsec:restoration_quantitative}

\begin{figure*}[!t]
\centering
\includegraphics[width=0.98\textwidth]{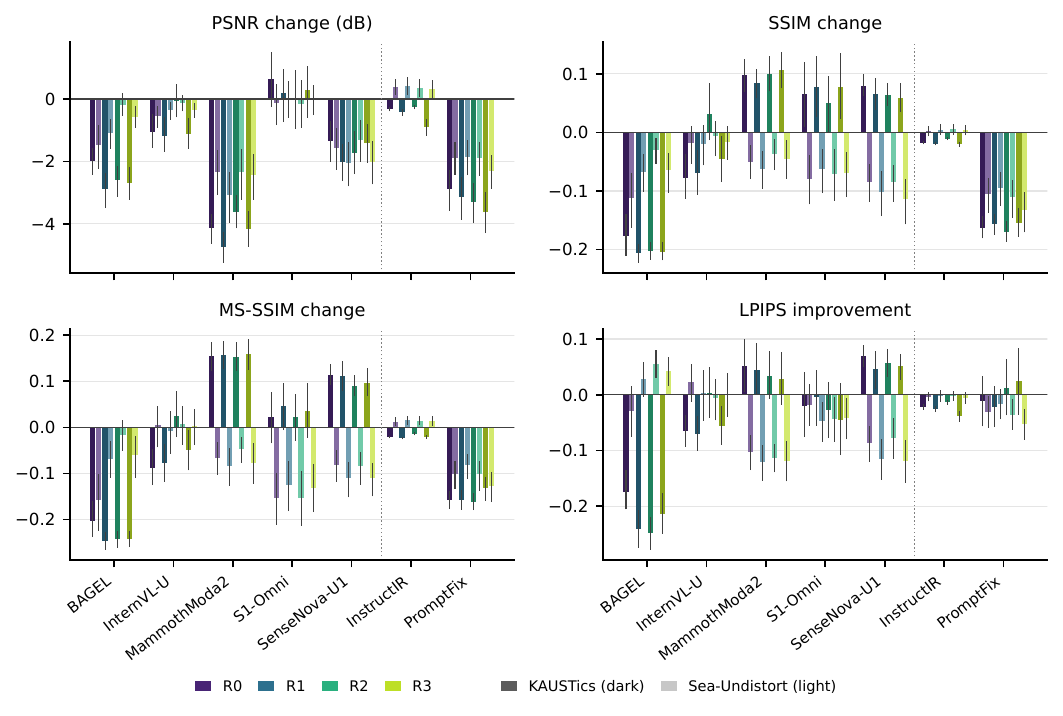}
\caption{Restoration change from the matching input baseline for R0--R3, for KAUSTics (dark) and Sea-Undistort (light). Error bars show paired image-bootstrap 95\% confidence intervals. Positive change indicates improvement. The dotted line separates the five unified MLLMs from InstructIR and PromptFix.}
\label{fig:restoration_changes}
\end{figure*}

Quantitative performance is summarized by prompt-specific changes in Fig.~\ref{fig:restoration_changes} and dataset-specific results at each metric's best pooled prompt in Table~\ref{tab:restoration_metrics}.

The restoration results provide little evidence that the zero-shot restoration stage reliably improves the distorted imagery. S1-Omni-Image reached a PSNR improvement of $+0.255$~dB under R0. InternVL-U remained close to its input in PSNR ($-0.096$~dB under R2), with small SSIM ($+0.013$) and MS-SSIM ($+0.017$) gains. MammothModa2 R2 improved SSIM by $+0.031$ and MS-SSIM by $+0.053$, but reduced PSNR by $2.980$~dB. BAGEL and SenseNova-U1 showed no pooled PSNR improvement.

The restoration-specific baselines also showed limited transfer. InstructIR reached $+0.040$~dB PSNR under R2, but its best SSIM change remained negative ($-0.003$). PromptFix worsened all four pooled metrics under every prompt, with best changes of $-2.394$~dB PSNR, $-0.126$ SSIM, $-0.120$ MS-SSIM, and $-0.012$ LPIPS improvement.

The small positive PSNR estimates were not clearly separated from zero: S1-Omni-Image R0 had a 95\% bootstrap interval of $[-0.310,\,0.829]$~dB, and InstructIR R2 had $[-0.120,\,0.206]$~dB. Identical inference repeats do not increase the number of independent images.

Dataset stratification exposed gains hidden by pooling. MammothModa2 R2 improved SSIM by $+0.099$ on KAUSTics (95\% CI $[0.070,0.131]$) but reduced it by $0.037$ on Sea-Undistort ($[-0.064,-0.011]$); PSNR declined on both datasets. InstructIR R2 changed PSNR by $-0.263$~dB on KAUSTics ($[-0.337,-0.196]$) but $+0.343$~dB on Sea-Undistort ($[0.068,0.645]$). Pooled performance therefore does not establish uniform behavior across semi-synthetic UAV and fully synthetic scenes.

The pooled LPIPS changes were negative for every model and prompt. Even when the best prompt is selected independently for each metric in Table~\ref{tab:restoration_metrics}, the best LPIPS change remains negative, ranging from $-0.001$ for InternVL-U to $-0.086$ for BAGEL. The table therefore provides a descriptive best-case summary.

Taken together, the quantitative results do not support treating any of the evaluated models as a reliable zero-shot restoration method for quantitative shallow-water remote sensing. Small improvements in individual metrics occur for particular model--prompt combinations, but they are not consistent across metrics, and the positive PSNR estimates for S1-Omni-Image and InstructIR are not clearly separated from zero under image-level bootstrap resampling. The results therefore indicate that successful degradation assessment does not, by itself, imply a reliable ability to reconstruct the underlying scene.

\begin{table*}[!t]
\centering
\caption{Dataset-specific restoration results for KAUSTics (K) and Sea-Undistort (S), with image counts in parentheses. For each metric, the best R0--R3 prompt is fixed. Each cell reports prompt, mean metric value, and the signed improvement over the matching input baseline in brackets. Bold marks the best value within each dataset. The last two models are restoration-only baselines.}
\label{tab:restoration_metrics}
\footnotesize
\setlength{\tabcolsep}{3pt}
\begin{adjustbox}{max width=\textwidth}
\begin{tabular}{llcccc}
\toprule
\textbf{Model} & \textbf{Dataset} & \textbf{PSNR (dB) $\uparrow$} & \textbf{SSIM $\uparrow$} & \textbf{MS-SSIM $\uparrow$} & \textbf{LPIPS $\downarrow$} \\
\midrule
BAGEL \cite{deng2025bagel} & K (50) & R2: 12.348 [$-2.597$] & R2: 0.114 [$-0.203$] & R2: 0.155 [$-0.243$] & R3: 0.924 [$-0.214$] \\
 & S (50) & R2: 10.374 [$-0.192$] & R2: 0.254 [$-0.031$] & R2: 0.414 [$-0.017$] & R3: \textbf{0.625} [$+0.042$] \\
\addlinespace[2pt]
InternVL-U \cite{tian2026internvlu} & K (50) & R2: 14.879 [$-0.065$] & R2: 0.349 [$+0.032$] & R2: 0.423 [$+0.026$] & R2: 0.706 [$+0.003$] \\
 & S (50) & R2: 10.439 [$-0.128$] & R2: 0.278 [$-0.007$] & R2: 0.439 [$+0.008$] & R2: 0.672 [$-0.005$] \\
\addlinespace[2pt]
MammothModa2 \cite{shen2025mammothmoda2} & K (50) & R2: 11.331 [$-3.614$] & R2: \textbf{0.416} [$+0.099$] & R2: \textbf{0.551} [$+0.153$] & R0: 0.659 [$+0.051$] \\
 & S (50) & R2: 8.221 [$-2.345$] & R2: 0.248 [$-0.037$] & R2: 0.384 [$-0.048$] & R0: 0.771 [$-0.104$] \\
\addlinespace[2pt]
S1-Omni-Image \cite{li2026s1omniimage} & K (50) & R0: \textbf{15.594} [$+0.650$] & R1: 0.395 [$+0.078$] & R1: 0.443 [$+0.045$] & R0: 0.729 [$-0.020$] \\
 & S (50) & R0: 10.428 [$-0.139$] & R1: 0.223 [$-0.062$] & R1: 0.305 [$-0.126$] & R0: 0.685 [$-0.018$] \\
\addlinespace[2pt]
SenseNova-U1 \cite{sensenova2026u1} & K (50) & R0: 13.595 [$-1.349$] & R0: 0.397 [$+0.080$] & R0: 0.512 [$+0.114$] & R0: \textbf{0.641} [$+0.069$] \\
 & S (50) & R0: 8.979 [$-1.587$] & R0: 0.200 [$-0.085$] & R0: 0.349 [$-0.082$] & R0: 0.755 [$-0.087$] \\
\addlinespace[2pt]
\midrule
InstructIR \cite{conde2024instructir} & K (50) & R2: 14.682 [$-0.263$] & R2: 0.307 [$-0.011$] & R2: 0.383 [$-0.014$] & R2: 0.723 [$-0.014$] \\
 & S (50) & R2: \textbf{10.909} [$+0.343$] & R2: \textbf{0.290} [$+0.005$] & R2: \textbf{0.445} [$+0.014$] & R2: 0.670 [$-0.003$] \\
\addlinespace[2pt]
PromptFix \cite{yu2024promptfix} & K (50) & R0: 12.063 [$-2.882$] & R1: 0.160 [$-0.157$] & R1: 0.239 [$-0.158$] & R2: 0.698 [$+0.012$] \\
 & S (50) & R0: 8.660 [$-1.906$] & R1: 0.190 [$-0.094$] & R1: 0.349 [$-0.082$] & R2: 0.703 [$-0.036$] \\
\bottomrule
\end{tabular}
\end{adjustbox}
\end{table*}

\subsubsection{Qualitative Behavior and Failure Modes}
\label{subsec:restoration_qualitative}

\begin{figure*}[!t]
\centering
\includegraphics[width=0.98\textwidth]{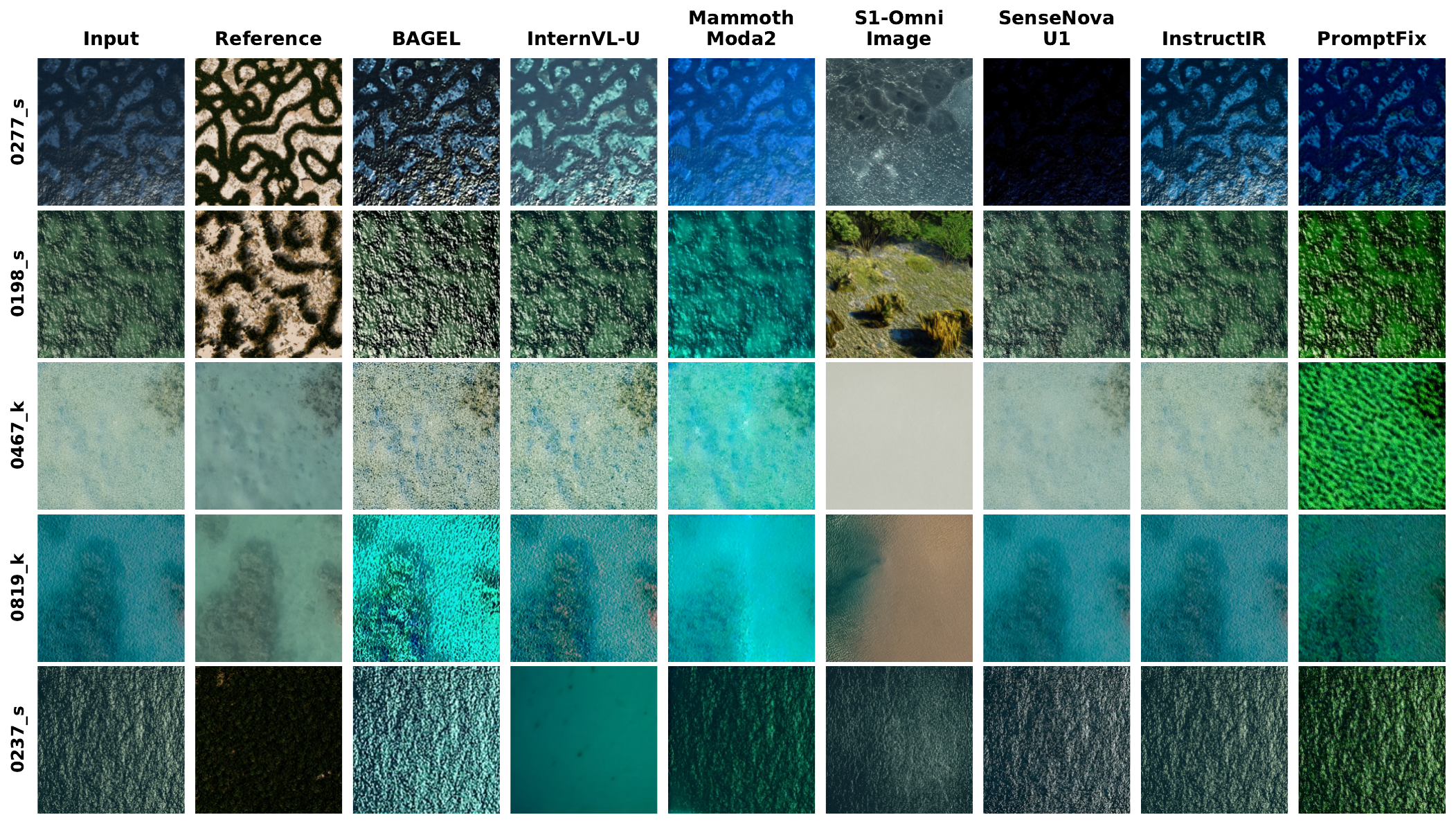}
\caption{Five common restoration examples using restoration prompt R0. Columns show the distorted input, paired reference, five unified MLLMs, InstructIR, and PromptFix. Suffixes \texttt{\_k} and \texttt{\_s} identify KAUSTics and Sea-Undistort datasets.The paired reference is displayed for comparison.}
\label{fig:seven_model_restoration}
\end{figure*}

Qualitative comparisons reveal changes in scene content that aggregate similarity metrics cannot fully describe. The five common scenes in Fig.~\ref{fig:seven_model_restoration} show, with prompt R0 fixed across all seven models, changes in color, contrast, texture, and geometry. Some outputs preserve the approximate spatial layout while modifying radiometry; others smooth or replace existing structure. Neither behavior consistently approaches the paired reference.



\begin{figure*}[!t]
\centering
\includegraphics[width=0.98\textwidth]{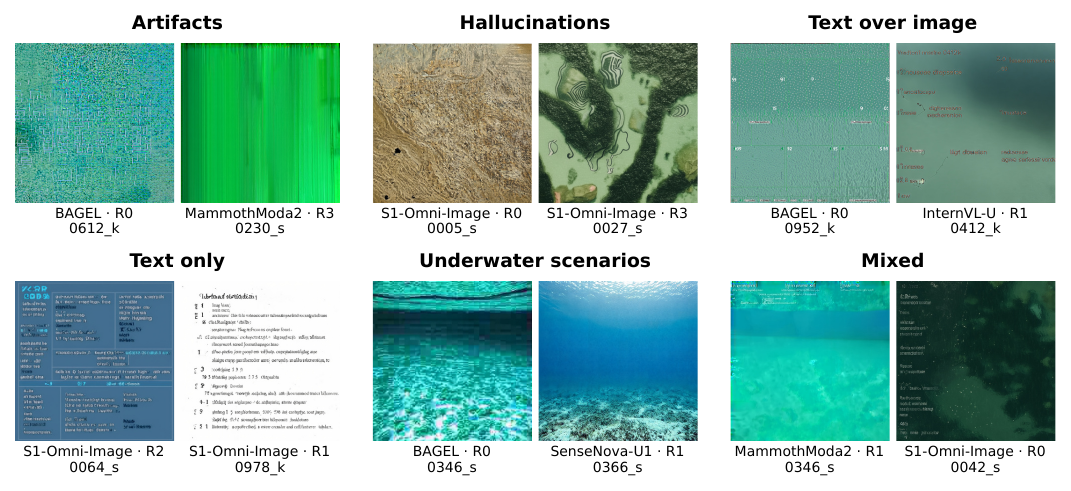}
\caption{Representative restoration failures grouped by visible behavior: artifacts, hallucinations, text over image, text only, underwater scenarios, and mixed failures. Two examples per category are shown. Labels identify the model, restoration prompt, and image. Text-only examples are dominated by document or interface content; underwater scenarios depict a generated submerged viewpoint. Categories can overlap.}
\label{fig:restoration_failure_gallery}
\end{figure*}

Average metric values can also conceal isolated but substantial generation failures. Severe failures fall into six descriptive groups: artifacts, hallucinations, text over image, text only, underwater scenarios, and mixed failures (Fig.~\ref{fig:restoration_failure_gallery}). Two manually selected examples illustrate each group. Categories can overlap, and their displayed counts do not estimate failure prevalence. Artifacts include repetitive textures, uniform regions, and abrupt boundaries. Hallucinations replace scene content; underwater scenarios specifically replace the UAV view with a submerged viewpoint. Text-over-image and text-only failures respectively overlay or replace the scene with text, while mixed failures combine several behaviors. None of the restoration prompts requests text or a changed viewpoint. These outputs introduce unsupported content rather than merely changing the degree of correction, limiting their suitability for quantitative remote sensing.

\subsubsection{Downstream Remote-Sensing Diagnostics}
\label{subsec:downstream_results}

\begin{figure*}[!t]
\centering
\includegraphics[width=0.98\textwidth]{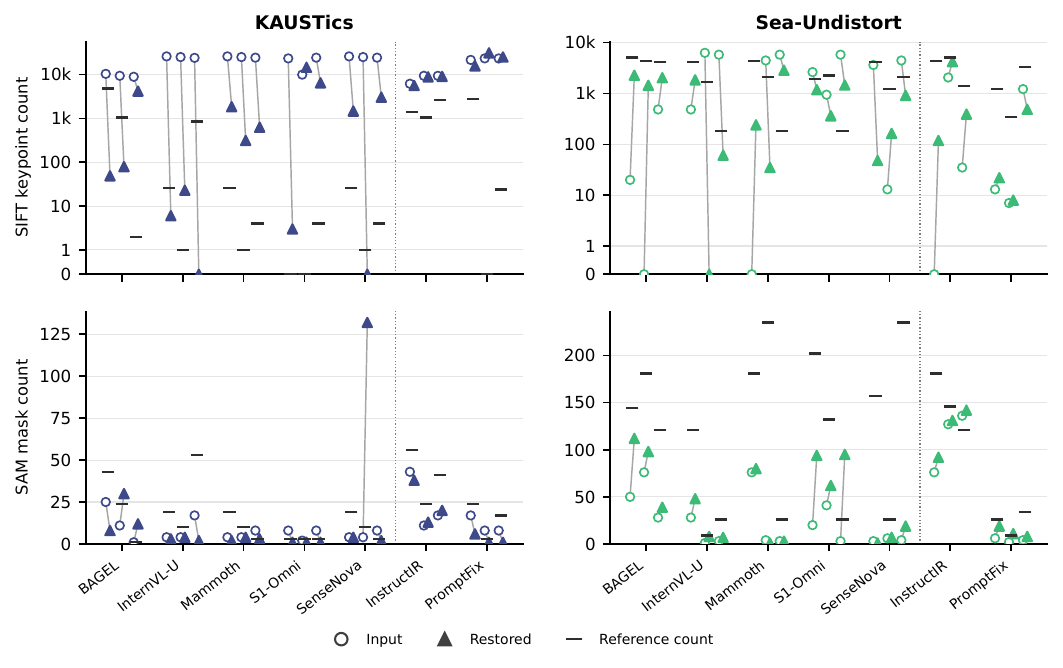}
\caption{Best-case downstream diagnostics for each model's own three highest-SSIM-gain restorations per dataset. Each group contains three input/restoration pairs; short black marks show the corresponding reference counts. The SIFT count axis is logarithmic above one and linear near zero. Moving the restored value closer to the reference indicates improved count agreement, but does not by itself establish recovery of the same physical features or regions.}
\label{fig:downstream_diagnostics}
\end{figure*}

Remote-sensing workflows require not only visually plausible imagery but also the preservation of stable spatial features that can support operations such as image matching, geometric reconstruction, and segmentation. To examine whether the apparent improvements observed in the full-reference metrics translate into more useful image structure, we additionally evaluated SIFT keypoints and SAM masks on the selected best-case restorations shown in Fig.~\ref{fig:downstream_diagnostics}.

For SIFT, the keypoint-count error relative to the paired reference decreased in approximately 86\% of the selected KAUSTics cases and 81\% of the Sea-Undistort cases. Thus, restoration frequently moved the total number of detected local features closer to that of the corresponding reference image. This result alone, however, does not establish that the correct physical features were recovered. A generated image may contain a similar number of keypoints to the reference while placing them on different textures, boundaries, or artificially generated structures. Conversely, excessive smoothing may remove distortion-induced features together with genuine seabed features.

A similar pattern was observed for SAM, for which the mask-count error decreased in approximately 38\% of the selected KAUSTics cases and 76\% of the Sea-Undistort cases. This suggests that restoration sometimes produced a more reference-like amount of segmented structure, particularly for Sea-Undistort. However, SAM mask counts are only a structural proxy: without semantic ground truth, closer counts do not establish recovery of the correct benthic objects or boundaries.

\begin{figure*}[!t]
\centering
\includegraphics[width=0.99\textwidth]{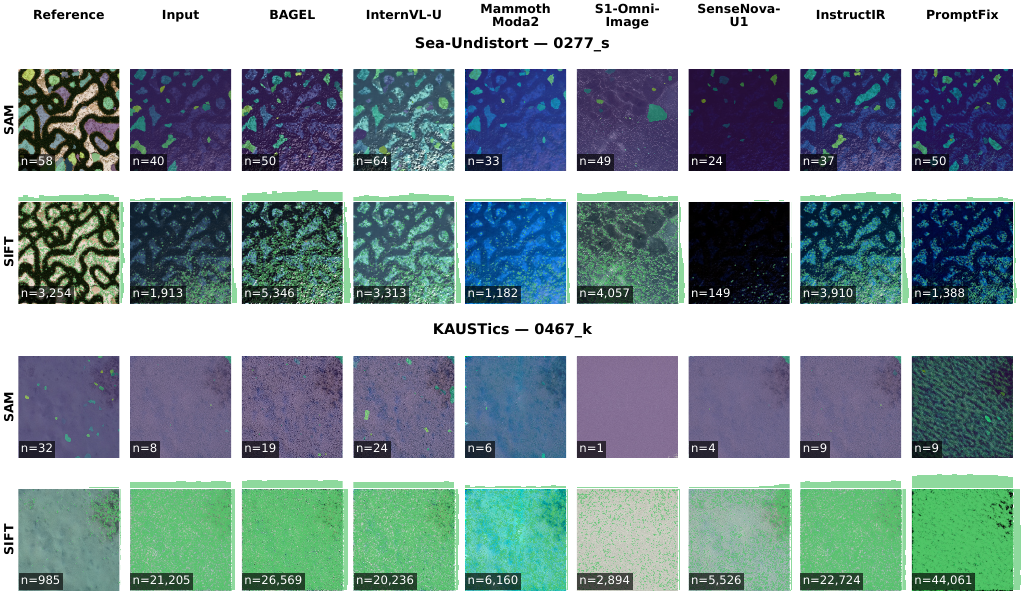}
\caption{Matched-image downstream diagnostics for Sea-Undistort 0277\_s and KAUSTics 0467\_k. Columns show the common reference, distorted input, and R0 restoration prompt from each model. The SAM row shows mask overlays and the SIFT row shows all detected keypoints together with top/right spatial histograms. Insets report $n$, the number of masks or keypoints. Histograms use 20 bins. Viridis mask colors distinguish regions locally and do not represent semantic classes. Similar feature or mask counts therefore do not necessarily indicate recovery of the same reference structure.}
\label{fig:downstream_matched}
\end{figure*}

The matched examples in Fig.~\ref{fig:downstream_matched} provide a complementary visual comparison by holding the scene and restoration prompt fixed across models. The same two scenes, Sea-Undistort 0277\_s and KAUSTics 0467\_k, are shown using prompt R0. The models produce substantially different SIFT keypoint densities and spatial distributions, as well as different SAM region decompositions. In some cases, restoration increases the number of detected features substantially without reproducing the spatial distribution visible in the reference; in others, the image becomes smoother and produces considerably fewer detectable structures. The SAM overlays similarly show that comparable mask counts can correspond to visibly different partitions of the scene.

Taken together, the SIFT and SAM diagnostics reinforce the distinction between improving image-level similarity and recovering scientifically useful scene structure. A restoration can move the number of detected keypoints or segmented regions toward the reference without necessarily recovering the same physical features or boundaries. The present analysis therefore does not demonstrate improved habitat classification or Structure-from-Motion performance. Rather, it shows that favorable full-reference metrics and more reference-like structural statistics should not automatically be interpreted as preservation of the underlying physical scene.

\section{Discussion}
\label{sec:discussion}

The unified MLLMs can recognize several visual signatures of shallow-water degradation and describe them using appropriate domain language, yet this perceptual capability does not translate into reliable reconstruction of the underlying scene. This distinction has important implications for the use of generative foundation models in quantitative remote sensing.

\subsection{The Gap Between Distortion Assessment and Restoration}
\label{subsec:discussion_assessment_restoration}

The assessment results show that pretrained unified MLLMs contain useful visual knowledge about shallow-water degradation. Several models can distinguish distortion types, estimate their severity levels, and provide short descriptions of the corresponding image evidence. However, identifying a distortion is fundamentally easier than reversing the physical process that produced it. Caustics can be recognized from characteristic illumination patterns, sun glint from bright specular regions, and turbidity from reduced visibility, but their correction requires estimating how the measured signal was altered between the seabed and the sensor.

This distinction is especially important because shallow-water restoration is an inverse problem. Refraction changes the apparent geometry of submerged features and cannot in general be reversed from appearance alone without information about the imaging geometry and water surface. Turbidity and light absorption reduce or redistribute the signal through scattering and attenuation, while strong sun glint can obscure or saturate seabed information. In these cases, a model may correctly recognize the distortion type while lacking sufficient information to recover what was physically present before degradation. Conventional restoration methods address this problem through explicit image-formation models, paired supervision, or physically simulated degradations \cite{akkaynak2019seathru,yue2023resshift,zhang2024underwaterdiffusion}. The unified MLLMs evaluated here receive none of this task-specific information.

The restoration results are consistent with this interpretation. Improvements depend on the metric, prompt, and dataset. No unified MLLM improves pooled LPIPS, and the positive pooled PSNR changes are small relative to their image-bootstrap intervals, although stratification reveals dataset-specific exceptions. The restoration-only baselines also transfer poorly under the tested shallow-water conditioning. This separation between understanding and generation is consistent with recent general-purpose unified-model benchmarks, which also report that strong understanding capabilities do not necessarily translate into equally reliable reasoning-guided generation \cite{shi2026realunify,li2026girbench}.

\subsection{Reliability in Scientific Remote-Sensing Workflows}
\label{subsec:discussion_reliability}

The qualitative failures highlight a broader concern: generative models can produce visually plausible outputs while no longer preserving the measured scene. Examples include synthetic textures, altered scene structure, generated underwater viewpoints, and unsupported text. Such hallucinations and failures of visual grounding are known limitations of large vision-language models \cite{li2023objecthallucination,guan2024hallusionbench}. In remote sensing, however, this is particularly problematic because radiometry, geometry, and spatial texture may themselves constitute the scientific information of interest. A visually cleaner image is therefore not necessarily a more accurate observation.

Prompt sensitivity introduces a related reproducibility issue. Changing only the assessment instruction altered presence balanced accuracy by up to 0.062 and the joint ordinal score by up to 0.103, while additional prompt structure did not consistently improve performance. Although repeated runs with identical settings produced identical outputs, changing the wording of the instruction altered model behavior. The models are therefore repeatable under a fixed protocol but not necessarily robust to prompt formulation, implying that operational use would require fixed, version-controlled prompts and revalidation when prompts or checkpoints change \cite{sahoo2024prompt,ismithdeen2025promptception}.

Human disagreement further emphasizes the difficulty of the task. Turbidity and refraction were more ambiguous for the annotators than caustics or sun glint, supporting the use of multiple expert ratings and disagreement-aware metrics. The evidence analysis also cautions against interpreting linguistic hedging as calibrated uncertainty, since hedge terms were associated more strongly with predicted severity than with correctness \cite{yona2024uncertainty}.

\subsection{Toward Physics-Informed Multimodal Restoration}
\label{subsec:discussion_physics}

The assessment capability observed here nevertheless suggests a useful role for multimodal models. Rather than solving the complete inverse problem directly, future systems could use MLLMs to identify distortions, estimate their severity, and determine the required corrections, with the resulting diagnosis conditioning a specialized restoration model or physics-based operator. Recent degradation-aware underwater enhancement methods provide evidence that explicitly conditioning restoration on degradation information can be beneficial \cite{yao2026conditional,wang2026degradation}. Related restoration agents already use multimodal models to diagnose degradations and orchestrate external tools \cite{chen2024restoreagent,zhou2025qagent}; shallow-water applications could extend this approach with operators tailored to refraction, glint, scattering, and wavelength-dependent attenuation.

A complementary direction is to adapt the unified models themselves. Parameter-efficient fine-tuning such as LoRA \cite{hu2022lora} on paired distorted/reference imagery could help learn the missing inverse mapping, while physical constraints or additional conditioning variables such as water depth and viewing geometry could further guide restoration. Structured conditioning of the predicted severity and confidence may also provide a more direct connection between diagnosis and generation than the current natural-language correction plan.

\subsection{Limitations and Operational Outlook}
\label{subsec:discussion_limitations}

Several limitations define the scope of these conclusions. The benchmark combines semi-synthetic distortions applied to real UAV imagery with fully synthetic through-water scenes, enabling paired full-reference evaluation but not capturing the full variability of naturally distorted field imagery. A large paired set of real degraded UAV observations with corresponding distortion-free references is not available, and such datasets are inherently difficult to construct because the same submerged scene cannot generally be observed simultaneously with and without water-surface and water-column effects. Restoration is also evaluated on a 100-image subset, while the two restoration-only baselines use fixed InternVL-U conditioning and model-specific preprocessing; their results should therefore not be interpreted as a general comparison with all specialized restoration methods.

The evaluation metrics introduce additional limitations. PSNR, SSIM, MS-SSIM, and LPIPS quantify similarity to the paired reference but do not establish physical fidelity or preservation of scientifically relevant information. The SIFT/SAM analysis provides only a limited best-case structural diagnostic and does not support claims about operational downstream accuracy. Future work should therefore include human fidelity assessment, distortion-specific measures, and full-cohort downstream tasks such as habitat mapping or geometric reconstruction. Overall, the present results support unified MLLMs more strongly as diagnostic and decision-support tools than as autonomous restoration systems, with greater near-term potential in human-in-the-loop workflows where physically constrained methods remain responsible for reconstruction.

\section{Conclusion}
\label{sec:conclusion}

This work presented a systematic benchmark of five open-weight unified MLLMs for the assessment and restoration of shallow-water UAV imagery. The benchmark contains 1,000 images, comprising 600 semi-synthetic distortions applied to real UAV imagery and 400 fully synthetic through-water scenes, with three independent expert annotations of five water-induced distortion types. All models were evaluated strictly zero-shot, using frozen pretrained weights and without access to paired reference images during either assessment or restoration. The assessment results show that unified MLLMs can recognize shallow-water degradation to a meaningful extent, although performance depends strongly on both the model and the prompt. At each model's best pooled presence-BA prompt, the highest presence BA was 0.654 and the highest joint ordinal score was 0.714. Excluding dataset-confounded caustics reduced the highest BA to 0.600 without changing the leading model; dataset-specific scores nevertheless show that pooled results do not establish uniform within-domain performance. These best-of-four benchmark comparisons are descriptive rather than held-out estimates. No single prompt was consistently best, and different models showed distinct prediction biases.

In the pooled analysis, restoration was considerably less reliable. Among the unified MLLMs, only S1-Omni-Image improved PSNR relative to the distorted input, with a maximum gain of $+0.255$~dB, while improvements in SSIM and MS-SSIM were generally small or accompanied by losses in pixel-level fidelity. The restoration-only baselines showed similarly limited transfer: InstructIR achieved a maximum PSNR gain of $+0.040$~dB, whereas PromptFix worsened all four metrics under every tested prompt. No model improved pooled LPIPS, and some generated outputs substantially altered scene structure or introduced unsupported content. Dataset stratification and the selected-case SIFT/SAM diagnostic further show that favorable similarity or count changes need not transfer across datasets or imply preservation of the underlying scene structure. These results indicate that current unified MLLMs are more promising as tools for diagnosing shallow-water image degradation than as reliable restoration engines for quantitative remote sensing. In the near term, their most useful role may therefore be in human-in-the-loop workflows, where they assist with identifying distortions, prioritizing corrections, or supporting restoration decisions, while physically constrained reconstruction remains handled by specialized methods. Future progress will likely depend on combining multimodal reasoning with task-specific adaptation, physical image-formation constraints, better uncertainty estimation, and validation based on downstream remote-sensing tasks. The proposed benchmark provides a reproducible baseline for measuring progress toward these goals.

\section*{Code Availability}
The source code, image IDs, annotation labels, complete prompts, JSON schema, evaluation code and model revision IDs for the proposed MLLMs framework are publicly available at \url{https://github.com/Dark3850/MLLMs_water}.

\section*{Acknowledgments}
The authors thank Hadeel A. Alamoudi of King Abdullah University of Science and Technology (KAUST) for labeling the data used in this study.


\ifCLASSOPTIONcaptionsoff
  \newpage
\fi



%

\phantomsection
\addcontentsline{toc}{section}{References}
\bibliographystyle{IEEEtran}
\bibliography{Bib/bibliography.bib}

\clearpage
\setcounter{section}{0}
\setcounter{figure}{0}
\setcounter{table}{0}
\renewcommand{\thesection}{S\arabic{section}}
\renewcommand{\thefigure}{S\arabic{figure}}
\renewcommand{\thetable}{S\arabic{table}}
\renewcommand{\theHsection}{supplementary.\arabic{section}}
\renewcommand{\theHfigure}{supplementary.\arabic{figure}}
\renewcommand{\theHtable}{supplementary.\arabic{table}}
\section*{Supplementary Material}
\input{Supplementary_Analyses}

\end{document}

%% file: Supplementary_Analyses.tex
\section{Evidence Length Across Models}
\label{app:evidence_length}

\begin{figure}[!h]
\centering
\includegraphics[width=\columnwidth]{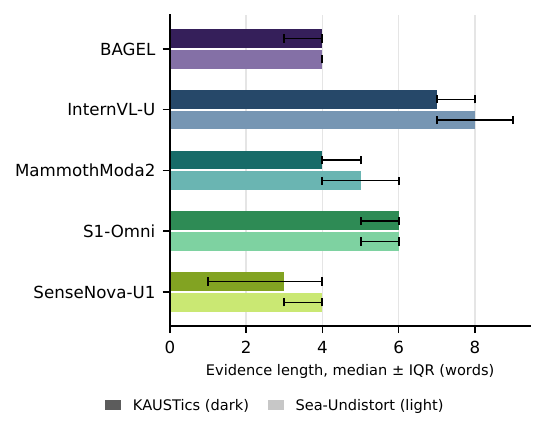}
\caption{Median evidence length by model and dataset, using all assessment evidence descriptions. Dark/light bars denote KAUSTics/Sea-Undistort; error bars show within-dataset interquartile ranges. The pooled comparison differs across models (Kruskal--Wallis $H=56{,}524$, $p<0.001$).}
\label{fig:word_count_by_model}
\end{figure}

The free-text evidence analysis in the main manuscript focuses primarily on the words used by the models. We additionally examined whether the length of these evidence descriptions differs systematically across models. Fig.~\ref{fig:word_count_by_model} reports the median number of words together with the interquartile range.

In the pooled evidence analysis, the models differ considerably in how much text they generate. Median evidence length ranges from 3 words for SenseNova-U1 to 7 words for InternVL-U, corresponding to more than a twofold difference. A Kruskal--Wallis test, which compares the distributions across the five models without assuming normality, confirms that these differences are statistically significant ($H=56{,}524$, $p<0.001$). Given the large number of evidence descriptions, however, the difference in median length is more informative than statistical significance alone. The result indicates that evidence length is strongly influenced by model-specific response style and should therefore not be interpreted directly as a measure of assessment quality or confidence.

\section{Severity Distribution Comparison}
\label{app:severity_distributions}

\begin{figure*}[!h]
\centering
\includegraphics[width=0.98\textwidth]{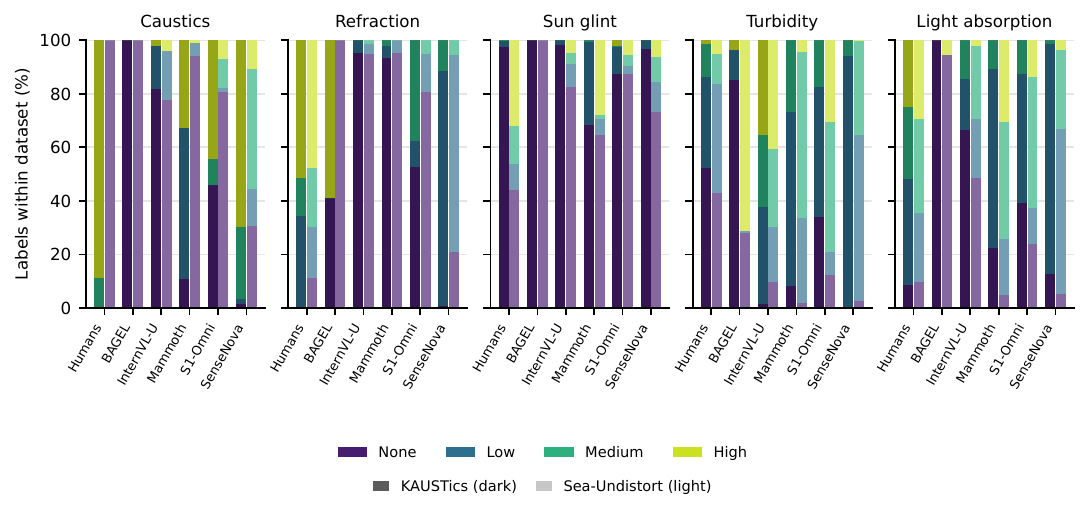}
\caption{Predicted severity distributions for each model's best pooled presence-BA prompt, as in Table~\ref{tab:assessment_results}, compared with the three human annotations and stratified by dataset. Dark/light bars denote KAUSTics/Sea-Undistort. Each bar shows within-dataset percentages assigned to \textit{none}, \textit{low}, \textit{medium}, and \textit{high}, and sums to 100\%.}
\label{fig:assessment_overall_severity}
\end{figure*}

The aggregate assessment metrics in Table~VI of the main manuscript summarize model performance with a small number of scores, but they do not show how frequently each severity level is predicted. Fig.~\ref{fig:assessment_overall_severity} therefore compares the predicted distributions of \textit{none}, \textit{low}, \textit{medium}, and \textit{high} with the combined labels of the three human annotators within each dataset and distortion.

Clear differences emerge across models. BAGEL predicts \textit{none} for most images and distortions, consistent with its low detection coverage reported in the model-level assessment results in the main manuscript. SenseNova-U1 shows the opposite tendency for several distortions, frequently predicting \textit{low} or \textit{medium}. MammothModa2 also frequently predicts \textit{low}, while the distributions vary substantially by distortion. Sun glint shows comparatively good agreement between model and human distributions, whereas turbidity remains difficult for all models. The latter result is consistent with the high human disagreement observed for turbidity in the human annotation results in the main manuscript.

\section{Severity Confusion Matrices by Distortion}
\label{app:confusion_matrices}

\begin{figure*}[!t]
\centering
\includegraphics[width=\textwidth]{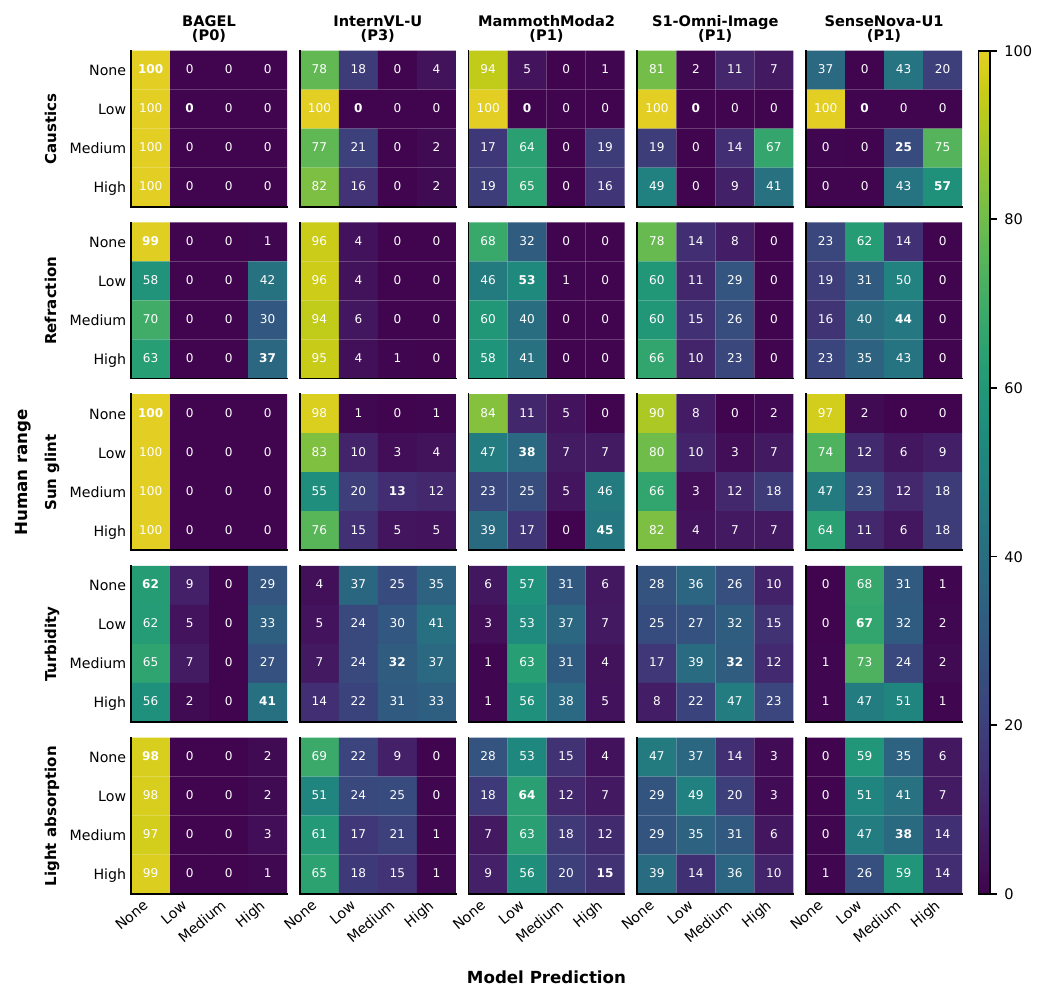}
\caption{Row-normalized severity confusion matrices for each model and distortion using that model's best pooled presence-BA prompt, as in Table~\ref{tab:assessment_results}. Rows represent the human severity grade and columns the model prediction. Each row sums to 100\%; stronger values along the diagonal indicate better severity discrimination, while a consistently dominant column indicates a systematic prediction bias. Bold diagonal values identify the highest correct-class rate across models for each distortion and human severity.}
\label{fig:assessment_confusion}
\end{figure*}

The severity distributions above show which classes each model tends to predict, but they do not indicate whether those predictions correspond to the human severity ratings. Fig.~\ref{fig:assessment_confusion} therefore reports a confusion matrix for every model and distortion. Each matrix is row-normalized: for a given human severity level, the values show the percentage of predictions assigned to each model severity level. A model with accurate severity discrimination would consequently concentrate its values along the diagonal.

The matrices reveal several model-specific biases. BAGEL strongly favors \textit{none}. For caustics and sun glint, 100\% of its predictions are \textit{none} regardless of the human severity, while approximately 97--99\% of its light-absorption predictions also fall in this class. InternVL-U shows a similar but less extreme tendency toward under-detection.

SenseNova-U1 displays a different behavior. Across human severity levels, approximately 47--73\% of its turbidity predictions and 26--59\% of its light-absorption predictions fall in the \textit{low} class, while \textit{none} is predicted in fewer than 2\% of cases. This tendency helps explain its high detection coverage but more moderate presence balanced accuracy. For caustics, approximately 25\% of \textit{medium} cases and 57\% of \textit{high} cases are correctly classified. MammothModa2 assigns \textit{low} to approximately 64--65\% of medium/high caustics cases, showing that strong presence detection need not imply accurate severity grading.

\section{Human Annotation Protocol}
\label{supp:annotation_protocol}

\subsection*{Annotators and Annotation Procedure}

The annotation panel comprised three Ph.D. students: two in Environmental Science and Engineering whose research involves remote sensing, and one in materials science/physics. Each annotator independently labeled all 1,000 distorted images. For every image, one of four ordered severity levels was assigned to each of the five distortion types. The interface required all five fields to be completed before a record could be saved, while an additional notes field was optional. The resulting three annotations per image were retained separately, without adjudication or consensus replacement.

The interface displayed the five distortion names and the four severity choices, but did not embed numerical thresholds or example images. Beyond identification of the categories and severity scale, no formal training or calibration set was used. Table~\ref{tab:supp_annotation_rubric} records the qualitative operational interpretation of the categories for reproducibility. Ratings therefore remained visual judgments, allowing genuine ambiguity to be represented by disagreement among annotators.

\begin{table*}[!ht]
\centering
\caption{Operational interpretation of the distortion and severity categories. No numerical thresholds or example anchors were embedded in the annotation interface.}
\label{tab:supp_annotation_rubric}
\small
\renewcommand{\arraystretch}{1.12}
\begin{tabular}{p{0.17\textwidth}p{0.77\textwidth}}
\toprule
\textbf{Category} & \textbf{Operational interpretation} \\
\midrule
Caustics & Focused or network-like bright patterns produced by surface-wave focusing; severity reflects their visual prominence rather than total scene brightness. \\
Refraction & Apparent spatial displacement, warping, or waviness of scene structure caused by refraction at the water surface. \\
Sun glint & Bright specular surface reflections, including saturated highlights that conceal underlying scene content. \\
Turbidity & Scattering-induced haze or veiling, reduced contrast, and loss of visibility attributable to suspended material. \\
Light absorption & Broad attenuation and wavelength-dependent loss of color or detail as light propagates through water. \\
\midrule
None & No visible evidence of the distortion. \\
Low & Subtle or localized evidence with little interference with scene interpretation. \\
Medium & Clearly visible or spatially extended evidence that moderately interferes with scene interpretation. \\
High & Dominant, widespread, or severe evidence that substantially obscures or alters scene content. \\
\bottomrule
\end{tabular}
\end{table*}

\subsection*{Presentation Conditions}

\begin{figure*}[!ht]
\centering
\includegraphics[width=0.96\textwidth]{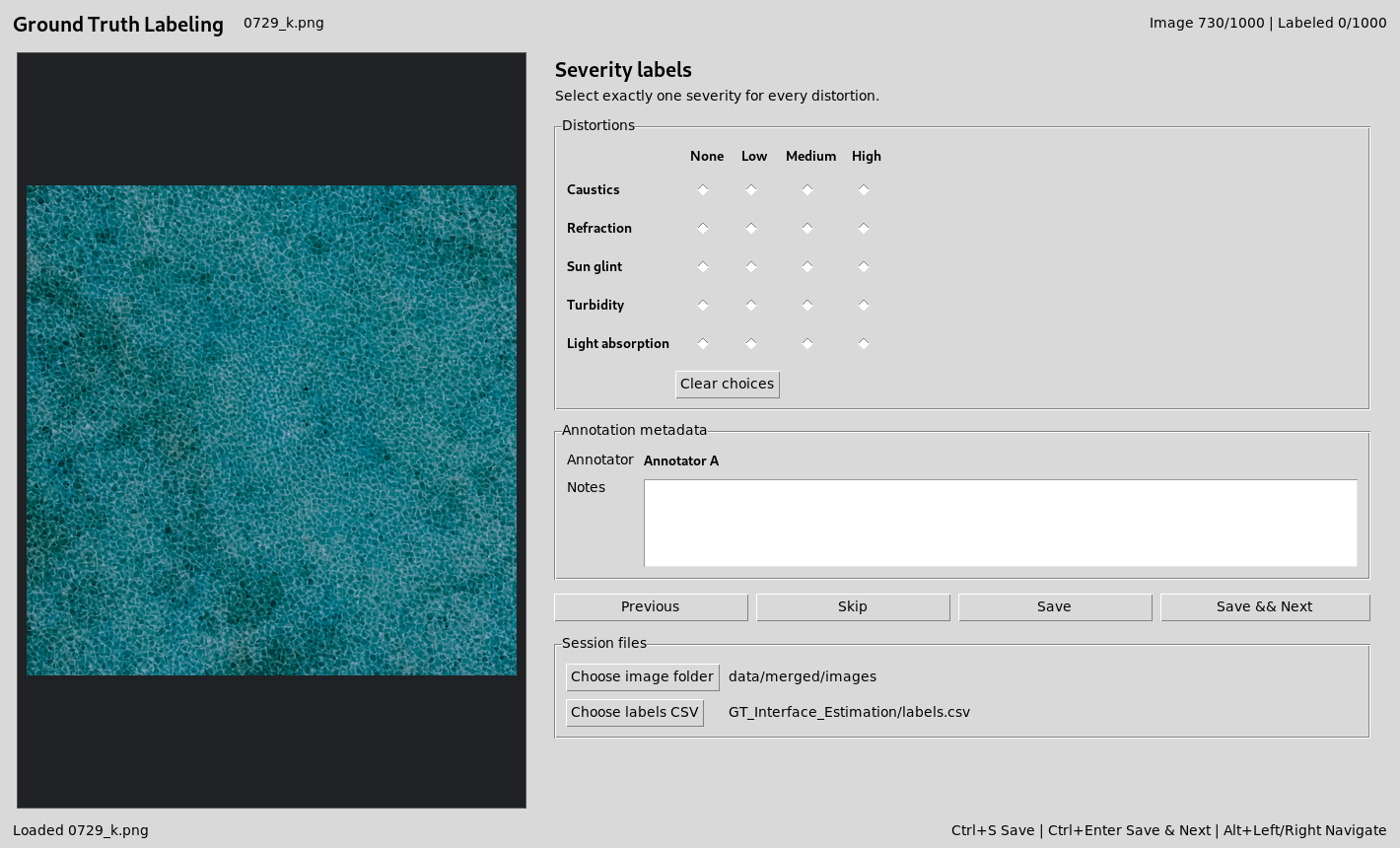}
\caption{Annotation interface used to present one distorted image and collect one mandatory severity selection for each distortion. The screenshot shows the unselected state; the notes field was optional.}
\label{fig:supp_annotation_interface}
\end{figure*}

Images were presented one at a time in a custom Python/Tkinter interface (Fig.~\ref{fig:supp_annotation_interface}). Only the distorted image was shown; annotators could not view the paired reference image or the other annotators' decisions. Images were traversed in a fixed lexicographic filename order rather than randomized. Relative filenames remained visible, and no additional procedure was used to conceal dataset or synthetic provenance. Images larger than the display canvas were downsampled for visualization using Lanczos resampling while preserving aspect ratio; the stored source images were unchanged.

Annotations were completed on the annotators' available displays. A common monitor model, color calibration, display luminance, viewing distance, and ambient-light condition were not imposed. These presentation conditions should therefore be considered when interpreting fine distinctions between adjacent severity levels.

%% file: Bib/bibliography.bib
@Inbook{Hobohm2021,
author="Hobohm, Carsten
and Schamin{\'e}e, Joop
and van Rooijen, Nils",
editor="Hobohm, Carsten",
title="Coastal Habitats, Shallow Seas and Inland Saline Steppes: Ecology, Distribution, Threats and Challenges",
bookTitle="Perspectives for Biodiversity and Ecosystems",
year="2021",
publisher="Springer International Publishing",
address="Cham",
pages="279--310",
isbn="978-3-030-57710-0",
doi="10.1007/978-3-030-57710-0_12",
url="https://doi.org/10.1007/978-3-030-57710-0_12"
}

@article{honda2013habitat,
    doi = {10.1371/journal.pone.0065735},
    author = {Honda, Kentaro AND Nakamura, Yohei AND Nakaoka, Masahiro AND Uy, Wilfredo H. AND Fortes, Miguel D.},
    journal = {PLOS ONE},
    publisher = {Public Library of Science},
    title = {Habitat Use by Fishes in Coral Reefs, Seagrass Beds and Mangrove Habitats in the Philippines},
    year = {2013},
    month = {08},
    volume = {8},
    url = {https://doi.org/10.1371/journal.pone.0065735},
    pages = {1-10},
    number = {8}
}

@article{guannel2016power,
    doi = {10.1371/journal.pone.0158094},
    author = {Guannel, Greg AND Arkema, Katie AND Ruggiero, Peter AND Verutes, Gregory},
    journal = {PLOS ONE},
    publisher = {Public Library of Science},
    title = {The Power of Three: Coral Reefs, Seagrasses and Mangroves Protect Coastal Regions and Increase Their Resilience},
    year = {2016},
    month = {07},
    volume = {11},
    url = {https://doi.org/10.1371/journal.pone.0158094},
    pages = {1-22},
    number = {7}
}

@article{kinsey1991significance,
title = {The significance of coral reefs as global carbon sinks— response to Greenhouse},
journal = {Palaeogeography, Palaeoclimatology, Palaeoecology},
volume = {89},
number = {4},
pages = {363-377},
year = {1991},
issn = {0031-0182},
doi = {https://doi.org/10.1016/0031-0182(91)90172-N},
url = {https://www.sciencedirect.com/science/article/pii/003101829190172N},
author = {D.W. Kinsey and D. Hopley}
}

@Article{gonzalez2020monitoring,
AUTHOR = {González-Rivero, Manuel and Beijbom, Oscar and Rodriguez-Ramirez, Alberto and Bryant, Dominic E. P. and Ganase, Anjani and Gonzalez-Marrero, Yeray and Herrera-Reveles, Ana and Kennedy, Emma V. and Kim, Catherine J. S. and Lopez-Marcano, Sebastian and Markey, Kathryn and Neal, Benjamin P. and Osborne, Kate and Reyes-Nivia, Catalina and Sampayo, Eugenia M. and Stolberg, Kristin and Taylor, Abbie and Vercelloni, Julie and Wyatt, Mathew and Hoegh-Guldberg, Ove},
TITLE = {Monitoring of Coral Reefs Using Artificial Intelligence: A Feasible and Cost-Effective Approach},
JOURNAL = {Remote Sensing},
VOLUME = {12},
YEAR = {2020},
NUMBER = {3},
ARTICLE-NUMBER = {489},
URL = {https://www.mdpi.com/2072-4292/12/3/489},
ISSN = {2072-4292}
}

@article{kroon2014informing,
title = {Informing policy to protect coastal coral reefs: Insight from a global review of reducing agricultural pollution to coastal ecosystems},
journal = {Marine Pollution Bulletin},
volume = {85},
number = {1},
pages = {33-41},
year = {2014},
issn = {0025-326X},
doi = {https://doi.org/10.1016/j.marpolbul.2014.06.003},
url = {https://www.sciencedirect.com/science/article/pii/S0025326X14003646},
author = {Frederieke J. Kroon and Britta Schaffelke and Rebecca Bartley}
}

@article{he2024remote,
title = {Remote sensing for shallow bathymetry: A systematic review},
journal = {Earth-Science Reviews},
volume = {258},
pages = {104957},
year = {2024},
issn = {0012-8252},
doi = {https://doi.org/10.1016/j.earscirev.2024.104957},
url = {https://www.sciencedirect.com/science/article/pii/S001282522400285X},
author = {Jinchen He and Shuhang Zhang and Xiaodong Cui and Wei Feng}
}

@article{kromer2025sea,
  title={Sea-Undistort: A Dataset for Through-Water Image Restoration in High Resolution Airborne Bathymetric Mapping},
  author={Kromer, Maximilian and Agrafiotis, Panagiotis and Demir, Beg{"u}m},
  journal={arXiv preprint arXiv:2508.07760},
  year={2025}
}

@ARTICLE{scilla2025computer,
  author={Scilla, Dario and Lopez, Omar A. and Nieuwenhuis, Brian Owain and Johansen, Kasper and Elías-Lara, Mariana and Angulo, Victor and Rodríguez, Jorge L. and Jones, Burton H. and McCabe, Matthew F.},
  journal={IEEE Journal of Selected Topics in Applied Earth Observations and Remote Sensing}, 
  title={Computer Vision Corrections Enhance UAV-Based Retrievals in Shallow Waters}, 
  year={2025},
  volume={18},
  number={},
  pages={18134-18149},
  doi={10.1109/JSTARS.2025.3587478},
  url = {https://ieeexplore.ieee.org/abstract/document/11075526}}

@article{tian2026internvlu,
  title={{InternVL-U}: Democratizing Unified Multimodal Models for Understanding, Reasoning, Generation and Editing},
  author={Tian, Changyao and Yang, Danni and Chen, Guanzhou and Cui, Erfei and Wang, Zhaokai and others},
  journal={arXiv preprint arXiv:2603.09877},
  year={2026}
}

@article{david_structure_motion_2021,
	title = {Structure-from-{Motion} on shallow reefs and beaches: potential and limitations of consumer-grade drones to reconstruct topography and bathymetry},
	volume = {40},
	issn = {1432-0975},
	shorttitle = {Structure-from-{Motion} on shallow reefs and beaches},
	url = {https://doi.org/10.1007/s00338-021-02088-9},
	doi = {10.1007/s00338-021-02088-9},
	language = {en},
	number = {3},
	urldate = {2026-08-23},
	journal = {Coral Reefs},
	author = {David, C. Gabriel and Kohl, Nina and Casella, Elisa and Rovere, Alessio and Ballesteros, Pablo and Schlurmann, Torsten},
	month = jun,
	year = {2021},
	pages = {835--851}
}

@article{ackleson2003light,
author = {Ackleson, Steven G.},
title = {Light in shallow waters: A brief research review},
journal = {Limnology and Oceanography},
volume = {48},
number = {1part2},
pages = {323-328},
year={2003},
doi = {https://doi.org/10.4319/lo.2003.48.1\_part\_2.0323},
}

@Article{agrafiotis2020correcting,
AUTHOR = {Agrafiotis, Panagiotis and Karantzalos, Konstantinos and Georgopoulos, Andreas and Skarlatos, Dimitrios},
TITLE = {Correcting Image Refraction: Towards Accurate Aerial Image-Based Bathymetry Mapping in Shallow Waters},
JOURNAL = {Remote Sensing},
VOLUME = {12},
YEAR = {2020},
NUMBER = {2},
ARTICLE-NUMBER = {322},
URL = {https://www.mdpi.com/2072-4292/12/2/322},
ISSN = {2072-4292},
DOI = {10.3390/rs12020322}
}

@Article{cui2022water,
AUTHOR = {Cui, Mengying and Sun, Yonghua and Huang, Chen and Li, Mengjun},
TITLE = {Water Turbidity Retrieval Based on UAV Hyperspectral Remote Sensing},
JOURNAL = {Water},
VOLUME = {14},
YEAR = {2022},
NUMBER = {1},
ARTICLE-NUMBER = {128},
URL = {https://www.mdpi.com/2073-4441/14/1/128},
ISSN = {2073-4441},
DOI = {10.3390/w14010128}
}

@misc{zhou2025qagent,
      title={Q-Agent: Quality-Driven Chain-of-Thought Image Restoration Agent through Robust Multimodal Large Language Model}, 
      author={Yingjie Zhou and Jiezhang Cao and Farong Wen and Zicheng Zhang and Yu Zhou and Yue Shi and Xiaohong Liu and Radu Timofte and Luc Van Gool and Guangtao Zhai},
      year={2026},
      eprint={2504.07148},
      archivePrefix={arXiv},
      primaryClass={eess.IV},
      url={https://arxiv.org/abs/2504.07148}, 
}

@misc{chen2024restoreagent,
      title={RestoreAgent: Autonomous Image Restoration Agent via Multimodal Large Language Models}, 
      author={Haoyu Chen and Wenbo Li and Jinjin Gu and Jingjing Ren and Sixiang Chen and Tian Ye and Renjing Pei and Kaiwen Zhou and Fenglong Song and Lei Zhu},
      year={2024},
      eprint={2407.18035},
      archivePrefix={arXiv},
      primaryClass={cs.CV},
      url={https://arxiv.org/abs/2407.18035}, 
}

@inproceedings{yona2024uncertainty,
    title = "Can Large Language Models Faithfully Express Their Intrinsic Uncertainty in Words?",
    author = "Yona, Gal  and
      Aharoni, Roee  and
      Geva, Mor",
    editor = "Al-Onaizan, Yaser  and
      Bansal, Mohit  and
      Chen, Yun-Nung",
    booktitle = "Proceedings of the 2024 Conference on Empirical Methods in Natural Language Processing",
    month = nov,
    year = "2024",
    address = "Miami, Florida, USA",
    publisher = "Association for Computational Linguistics",
    url = "https://aclanthology.org/2024.emnlp-main.443/",
    doi = "10.18653/v1/2024.emnlp-main.443",
    pages = "7752--7764"
}

@article{shen2025mammothmoda2,
  title={{MammothModa2}: A Unified {AR}-Diffusion Framework for Multimodal Understanding and Generation},
  author={Shen, Tao and Wan, Xin and Chen, Taicai and Zhang, Rui and Pan, Junwen and Lu, Dawei and Lei, Fanding and Lu, Zhilin and Yang, Yunfei and Cheng, Chen and She, Qi and Liu, Chang and Sun, Zhenbang},
  journal={arXiv preprint arXiv:2511.18262},
  year={2025}
}

@article{sensenova2026u1,
  title={{SenseNova-U1}: Unifying Multimodal Understanding and Generation with {NEO-unify} Architecture},
  author={{SenseNova Team}},
  journal={arXiv preprint arXiv:2605.12500},
  year={2026}
}

@article{deng2025bagel,
  title={Emerging Properties in Unified Multimodal Pretraining},
  author={Deng, Chaorui and Zhu, Deyao and Li, Kunchang and Gou, Chenhui and Li, Feng and Wang, Zeyu and Zhong, Shu and Yu, Weihao and Nie, Xiaonan and Song, Ziang and Shi, Guang and Fan, Haoqi},
  journal={arXiv preprint arXiv:2505.14683},
  year={2025}
}

@article{li2026s1omniimage,
  title={{S1-Omni-Image}: A Unified Model for Scientific Image Understanding, Generation, and Editing},
  author={Li, Qingxiao and Wang, Zikai and Wang, Qingli and Xu, Nan},
  journal={arXiv preprint arXiv:2606.24441},
  year={2026}
}

@article{cohen1968weighted,
  title={Weighted Kappa: Nominal Scale Agreement with Provision for Scaled Disagreement or Partial Credit},
  author={Cohen, Jacob},
  journal={Psychological Bulletin},
  volume={70},
  number={4},
  pages={213--220},
  year={1968},
  doi={10.1037/h0026256}
}

@article{gneiting2007scoring,
  title={Strictly Proper Scoring Rules, Prediction, and Estimation},
  author={Gneiting, Tilmann and Raftery, Adrian E.},
  journal={Journal of the American Statistical Association},
  volume={102},
  number={477},
  pages={359--378},
  year={2007},
  doi={10.1198/016214506000001437}
}

@book{efron1994bootstrap,
  title={An Introduction to the Bootstrap},
  author={Efron, Bradley and Tibshirani, Robert J.},
  publisher={Chapman and Hall/CRC},
  address={New York},
  year={1994}
}

@article{wang2004ssim,
  title={Image Quality Assessment: From Error Visibility to Structural Similarity},
  author={Wang, Zhou and Bovik, Alan C. and Sheikh, Hamid R. and Simoncelli, Eero P.},
  journal={IEEE Transactions on Image Processing},
  volume={13},
  number={4},
  pages={600--612},
  year={2004},
  doi={10.1109/TIP.2003.819861}
}

@inproceedings{wang2003msssim,
  title={Multiscale Structural Similarity for Image Quality Assessment},
  author={Wang, Zhou and Simoncelli, Eero P. and Bovik, Alan C.},
  booktitle={The Thirty-Seventh Asilomar Conference on Signals, Systems and Computers},
  volume={2},
  pages={1398--1402},
  year={2003},
  doi={10.1109/ACSSC.2003.1292216}
}

@inproceedings{zhang2018lpips,
  title={The Unreasonable Effectiveness of Deep Features as a Perceptual Metric},
  author={Zhang, Richard and Isola, Phillip and Efros, Alexei A. and Shechtman, Eli and Wang, Oliver},
  booktitle={Proceedings of the IEEE Conference on Computer Vision and Pattern Recognition},
  pages={586--595},
  year={2018}
}

@article{sahoo2024prompt,
  title={A Systematic Survey of Prompt Engineering in Large Language Models: Techniques and Applications},
  author={Sahoo, Pranab and Singh, Ayush Kumar and Saha, Sriparna and Jain, Vinija and Mondal, Samrat and Chadha, Aman},
  journal={arXiv preprint arXiv:2402.07927},
  year={2024},
  doi={10.48550/arXiv.2402.07927}
}

@inproceedings{wei2022chain,
  title={Chain-of-Thought Prompting Elicits Reasoning in Large Language Models},
  author={Wei, Jason and Wang, Xuezhi and Schuurmans, Dale and Bosma, Maarten and Ichter, Brian and Xia, Fei and Chi, Ed H. and Le, Quoc V. and Zhou, Denny},
  booktitle={Advances in Neural Information Processing Systems},
  volume={35},
  pages={24824--24837},
  year={2022}
}

@inproceedings{kojima2022large,
  title={Large Language Models are Zero-Shot Reasoners},
  author={Kojima, Takeshi and Gu, Shixiang Shane and Reid, Machel and Matsuo, Yutaka and Iwasawa, Yusuke},
  booktitle={Advances in Neural Information Processing Systems},
  volume={35},
  pages={22199--22213},
  year={2022}
}

@article{obura2019coral,
  title={Coral Reef Monitoring, Reef Assessment Technologies, and Ecosystem-Based Management},
  author={Obura, David O. and others},
  journal={Frontiers in Marine Science},
  volume={6},
  pages={580},
  year={2019},
  doi={10.3389/fmars.2019.00580}
}

@article{duplancicleder2023optical,
  title={Optical Satellite-Derived Bathymetry: An Overview and {WoS} and {S}copus Bibliometric Analysis},
  author={Duplan{\v{c}}i{\'c} Leder, Tea and Bau{\v{c}}i{\'c}, Martina and Leder, Nenad and Gili{\'c}, Frane},
  journal={Remote Sensing},
  volume={15},
  number={5},
  pages={1294},
  year={2023},
  doi={10.3390/rs15051294}
}

@article{hedley2005sunglint,
  title={Simple and Robust Removal of Sun Glint for Mapping Shallow-Water Benthos},
  author={Hedley, J. D. and Harborne, A. R. and Mumby, P. J.},
  journal={International Journal of Remote Sensing},
  volume={26},
  number={10},
  pages={2107--2112},
  year={2005},
  doi={10.1080/01431160500034086}
}

@article{he2011dark,
  title={Single Image Haze Removal Using Dark Channel Prior},
  author={He, Kaiming and Sun, Jian and Tang, Xiaoou},
  journal={IEEE Transactions on Pattern Analysis and Machine Intelligence},
  volume={33},
  number={12},
  pages={2341--2353},
  year={2011},
  doi={10.1109/TPAMI.2010.168}
}

@inproceedings{akkaynak2019seathru,
  title={Sea-thru: A Method for Removing Water from Underwater Images},
  author={Akkaynak, Derya and Treibitz, Tali},
  booktitle={Proceedings of the IEEE/CVF Conference on Computer Vision and Pattern Recognition},
  pages={1682--1691},
  year={2019},
  doi={10.1109/CVPR.2019.00178}
}

@article{anwar2020diving,
  title={Diving Deeper into Underwater Image Enhancement: A Survey},
  author={Anwar, Saeed and Li, Chongyi},
  journal={Signal Processing: Image Communication},
  volume={89},
  pages={115978},
  year={2020},
  doi={10.1016/j.image.2020.115978}
}

@article{islam2020funie,
  title={Fast Underwater Image Enhancement for Improved Visual Perception},
  author={Islam, Md Jahidul and Xia, Youya and Sattar, Junaed},
  journal={IEEE Robotics and Automation Letters},
  volume={5},
  number={2},
  pages={3227--3234},
  year={2020},
  doi={10.1109/LRA.2020.2974710}
}

@inproceedings{ho2020ddpm,
  title={Denoising Diffusion Probabilistic Models},
  author={Ho, Jonathan and Jain, Ajay and Abbeel, Pieter},
  booktitle={Advances in Neural Information Processing Systems},
  volume={33},
  pages={6840--6851},
  year={2020}
}

@inproceedings{rombach2022ldm,
  title={High-Resolution Image Synthesis with Latent Diffusion Models},
  author={Rombach, Robin and Blattmann, Andreas and Lorenz, Dominik and Esser, Patrick and Ommer, Bj{\"o}rn},
  booktitle={Proceedings of the IEEE/CVF Conference on Computer Vision and Pattern Recognition},
  pages={10674--10685},
  year={2022},
  doi={10.1109/CVPR52688.2022.01042}
}

@inproceedings{yue2023resshift,
  title={{ResShift}: Efficient Diffusion Model for Image Super-Resolution by Residual Shifting},
  author={Yue, Zongsheng and Wang, Jianyi and Loy, Chen Change},
  booktitle={Advances in Neural Information Processing Systems},
  volume={36},
  year={2023}
}

@article{zhang2024underwaterdiffusion,
  title={An Underwater Image Enhancement Method Based on Diffusion Model Using Dual-Layer Attention Mechanism},
  author={Zhang, Hong and He, Ran and Fang, Wei},
  journal={Water},
  volume={16},
  number={13},
  pages={1813},
  year={2024},
  doi={10.3390/w16131813}
}

@inproceedings{wu2024mllmiqa,
  title={A Comprehensive Study of Multimodal Large Language Models for Image Quality Assessment},
  author={Wu, Tianhe and Ma, Kede and Liang, Jie and Yang, Yujiu and Zhang, Lei},
  booktitle={Computer Vision -- ECCV 2024},
  series={Lecture Notes in Computer Science},
  year={2024},
  doi={10.1007/978-3-031-72904-1_9}
}

@article{dong2023icl,
  title={A Survey on In-Context Learning},
  author={Dong, Qingxiu and Li, Lei and Dai, Damai and Zheng, Ce and Ma, Jingyuan and Li, Rui and Xia, Heming and Xu, Jingjing and Wu, Zhiyong and Liu, Tianyu and Chang, Baobao and Sun, Xu and Li, Lei and Sui, Zhifang},
  journal={arXiv preprint arXiv:2301.00234},
  year={2023}
}

@inproceedings{liu2023llava,
  title={Visual Instruction Tuning},
  author={Liu, Haotian and Li, Chunyuan and Wu, Qingyang and Lee, Yong Jae},
  booktitle={Advances in Neural Information Processing Systems},
  volume={36},
  year={2023}
}

@techreport{openai2023gpt4v,
  title={{GPT-4V(ision)} System Card},
  author={{OpenAI}},
  institution={OpenAI},
  year={2023}
}

@article{ismithdeen2025promptception,
  title={Promptception: How Sensitive Are Large Multimodal Models to Prompts?},
  author={Ismithdeen, Mohamed Insaf and Khattak, Muhammad Uzair and Khan, Salman},
  journal={arXiv preprint arXiv:2509.03986},
  year={2025}
}

@inproceedings{conde2024instructir,
  title = {{InstructIR}: High-Quality Image Restoration Following Human Instructions},
  author = {Conde, Marcos V. and Geigle, Gregor and Timofte, Radu},
  booktitle = {European Conference on Computer Vision (ECCV)},
  year = {2024},
  url = {https://www.ecva.net/papers/eccv_2024/papers_ECCV/html/5238_ECCV_2024_paper.php}
}

@inproceedings{yu2024promptfix,
  title = {{PromptFix}: You Prompt and We Fix the Photo},
  author = {Yu, Yongsheng and Zeng, Ziyun and Hua, Hang and Fu, Jianlong and Luo, Jiebo},
  booktitle = {Advances in Neural Information Processing Systems},
  volume = {37},
  year = {2024},
  url = {https://proceedings.neurips.cc/paper_files/paper/2024/hash/468c9ec4215e05b6488ac307d377662e-Abstract-Conference.html}
}

@article{cao2026uq,
	title = {{UQ}-{Bench}: {A} {Benchmark} for {Evaluating} {Multimodal} {LLMs} on {Underwater} {Image} {Quality} {Assessment}},
	volume = {40},
	copyright = {Copyright (c) 2026 Association for the Advancement of Artificial Intelligence},
	issn = {2374-3468},
	shorttitle = {{UQ}-{Bench}},
	url = {https://ojs.aaai.org/index.php/AAAI/article/view/37246},
	doi = {10.1609/aaai.v40i4.37246},
	language = {en},
	number = {4},
	urldate = {2026-06-28},
	journal = {Proceedings of the AAAI Conference on Artificial Intelligence},
	author = {Cao, Jingchao and An, Guo and Gao, Feng and Gu, Ke and Liu, Yutao},
	month = mar,
	year = {2026},
	pages = {2589--2597}
}

@article{yao2026conditional,
	title = {Conditional {Prompt} {Learning} via {Degradation} {Perception} for {Underwater} {Image} {Enhancement}},
	volume = {40},
	copyright = {Copyright (c) 2026 Association for the Advancement of Artificial Intelligence},
	issn = {2374-3468},
	url = {https://ojs.aaai.org/index.php/AAAI/article/view/38176},
	doi = {10.1609/aaai.v40i14.38176},
	language = {en},
	number = {14},
	urldate = {2026-09-09},
	journal = {Proceedings of the AAAI Conference on Artificial Intelligence},
	author = {Yao, Mingze and Jiang, Zhiying and Fu, Xianping and Wang, Huibing},
	month = mar,
	year = {2026},
	pages = {11892--11900}
}

@inproceedings{li2023objecthallucination,
  title     = {Evaluating Object Hallucination in Large Vision-Language Models},
  author    = {Li, Yifan and Du, Yifan and Zhou, Kun and Wang, Jinpeng and Zhao, Xin and Wen, Ji-Rong},
  booktitle = {Proceedings of the 2023 Conference on Empirical Methods in Natural Language Processing},
  pages     = {292--305},
  year      = {2023},
  publisher = {Association for Computational Linguistics},
  doi       = {10.18653/v1/2023.emnlp-main.20}
}

@inproceedings{guan2024hallusionbench,
  title     = {HallusionBench: An Advanced Diagnostic Suite for Entangled Language Hallucination and Visual Illusion in Large Vision-Language Models},
  author    = {Guan, Tianrui and Liu, Fuxiao and Wu, Xiyang and Xian, Ruiqi and Li, Zongxia and Liu, Xiaoyu and Wang, Xijun and Chen, Lichang and Huang, Furong and Yacoob, Yaser and Manocha, Dinesh and Zhou, Tianyi},
  booktitle = {Proceedings of the IEEE/CVF Conference on Computer Vision and Pattern Recognition},
  pages     = {14375--14385},
  year      = {2024},
  doi       = {10.1109/CVPR52733.2024.01363}
}

@inproceedings{hu2022lora,
  title     = {LoRA: Low-Rank Adaptation of Large Language Models},
  author    = {Hu, Edward J. and Shen, Yelong and Wallis, Phillip and Allen-Zhu, Zeyuan and Li, Yuanzhi and Wang, Shean and Wang, Lu and Chen, Weizhu},
  booktitle = {International Conference on Learning Representations},
  year      = {2022}
}

@article{lowe2004sift,
  title = {Distinctive Image Features from Scale-Invariant Keypoints},
  author = {Lowe, David G.},
  journal = {International Journal of Computer Vision},
  year = {2004},
  url = {https://www.cs.ubc.ca/~lowe/papers/ijcv04.pdf}
}

@article{kirillov2023segany,
  title = {Segment Anything},
  author = {Kirillov, Alexander and Mintun, Eric and Ravi, Nikhila and Mao, Hanzi and Rolland, Chloe and Gustafson, Laura and Xiao, Tete and Whitehead, Spencer and Berg, Alexander C. and Lo, Wan-Yen and Doll{\'a}r, Piotr and Girshick, Ross},
  journal = {arXiv:2304.02643},
  year = {2023}
}

@inproceedings{shi2026realunify,
  title     = {RealUnify: Do Unified Models Truly Benefit from Unification? A Comprehensive Benchmark},
  author    = {Shi, Yang and Dong, Yuhao and Ding, Yue and Wang, Yuran and Zhu, Xuanyu and Zhou, Sheng and Liu, Wenting and Tian, Haochen and Wang, Rundong and Wang, Huanqian and Liu, Zuyan and Zeng, Bohan and Chen, Ruizhe and Wang, Qixun and Zhang, Zhuoran and Chen, Xinlong and Tong, Chengzhuo and Li, Bozhou and Liu, Qiang and Wang, Haotian and Yang, Wenjing and Zhang, Yuanxing and Wan, Pengfei and Zhang, Yi-Fan and Liu, Ziwei},
  booktitle = {Proceedings of the IEEE/CVF Conference on Computer Vision and Pattern Recognition},
  pages     = {22488--22497},
  year      = {2026}
}

@inproceedings{li2026girbench,
  title     = {{GIR-Bench}: Versatile Benchmark for Generating Images with Reasoning},
  author    = {Li, Hongxiang and Li, Yaowei and Lin, Bin and Niu, Yuwei and Yang, Yuhang and Huang, Xiaoshuang and Cai, Jiayin and Jiang, Xiaolong and Hu, Yao and Chen, Long},
  booktitle = {International Conference on Learning Representations},
  year      = {2026}
}

@inproceedings{xie2026mmeunify,
  title     = {{MME-Unify}: A Comprehensive Benchmark for Unified Multimodal Understanding and Generation Models},
  author    = {Xie, Wulin and Zhang, Yi-Fan and Fu, Chaoyou and Shi, Yang and Nie, Bingyan and Chen, Hongkai and Zhang, Zhang and Wang, Liang},
  booktitle = {International Conference on Learning Representations},
  year      = {2026}
}

@article{wang2026degradation,
  title   = {Underwater Image Enhancement via Degradation Information Extraction and Guidance},
  author  = {Wang, Fukuan and Li, Fei and Cen, Chaojun and Li, Zhenbo and Duan, Qingling},
  journal = {Pattern Recognition},
  volume  = {175},
  pages   = {113121},
  year    = {2026},
  month   = jul,
  doi     = {10.1016/j.patcog.2026.113121}
}

@misc{scilla2026kaustics,
      title={KAUSTics-Net: Diffusion-Based Through-Water UAV Restoration with a Semi-Synthetic Dataset}, 
      author={Dario Scilla and Omar A. Lopez and Kasper Johansen and Brian O. Nieuwenhuis and Matthew F. McCabe},
      year={2026},
      eprint={2611.03120},
      archivePrefix={arXiv},
      primaryClass={cs.CV},
      url={https://arxiv.org/abs/2611.03120}, 
}
